\documentclass[10pt,journal,compsoc,a4,twoside]{IEEEtran}
\ifCLASSOPTIONcompsoc
  \usepackage[nocompress]{cite}
\else
  \usepackage{cite}
\fi
\usepackage{url}
\usepackage[pagebackref=true,breaklinks=true,letterpaper=true,colorlinks,bookmarks=false]{hyperref}

\ifCLASSINFOpdf
\else
\fi

\usepackage[utf8]{inputenc}
\usepackage{graphicx}
\usepackage[fleqn]{amsmath}
\usepackage{amssymb}
\usepackage{subfigure}
\usepackage{multirow}
\usepackage{enumitem}
\usepackage{url}
\usepackage{float}
\usepackage{tabularx}
\usepackage{color}
\usepackage{soul}
\usepackage{algorithm, algorithmic}
\usepackage{braket}
\usepackage[table,x11names]{xcolor}
\usepackage{xcolor}
\soulregister\cite7
\soulregister\ref7
\newcommand{\minisection}[1]{\vspace{0.04in} \noindent {\bf #1}\ \ }

\iftrue

\else
\fi

\begin{document}
\bstctlcite{IEEEexample:BSTcontrol}   

\title{Exploiting Unlabeled Data in CNNs by Self-supervised Learning to Rank}

\author{Xialei Liu,
Joost van de Weijer, 
and Andrew D. Bagdanov
\IEEEcompsocitemizethanks{\IEEEcompsocthanksitem X. Liu and J. van de
  Weijer are with the Computer Vision Center at the Universitat Autònoma de Barcelona.\protect\\
E-mail: {\tt xialei@cvc.uab.es, joost@cvc.uab.es}
\IEEEcompsocthanksitem A. D. Bagdanov is with the Media Integration
and Communication Center, at the University of Florence.
E-mail: {\tt andrew.bagdanov@unifi.it}
}

\thanks{}}

\markboth{IEEE Transactions on Pattern Analysis and Machine Intelligence}{Liu, van de Weijer, and Bagdanov: Exploiting Unlabeled Data in CNNs by Self-supervised Learning to Rank}

\IEEEtitleabstractindextext{%
\begin{abstract}
  For many applications the collection of labeled data is expensive
  laborious. Exploitation of unlabeled data during training is thus a
  long pursued objective of machine learning. Self-supervised learning
  addresses this by positing an auxiliary task (different, but related
  to the supervised task) for which data is abundantly available. In
  this paper, we show how ranking can be used as a proxy task for some
  regression problems. As another contribution, we propose an
  efficient backpropagation technique for Siamese networks which
  prevents the redundant computation introduced by the multi-branch
  network architecture.

  We apply our framework to two regression problems: Image Quality
  Assessment (IQA) and Crowd Counting. For both we show how to
  automatically generate ranked image sets from unlabeled data. Our
  results show that networks trained to regress to the ground truth
  targets for labeled data and to simultaneously learn to rank
  unlabeled data obtain significantly better, state-of-the-art results
  for both IQA and crowd counting. In addition, we show that measuring
  network uncertainty on the self-supervised proxy task is a good
  measure of informativeness of unlabeled data. This can be
    used to drive an algorithm for active learning and we show that
    this reduces labeling effort by up to 50\%.
\end{abstract}

\begin{IEEEkeywords}
Learning from rankings, image quality assessment, crowd counting, active learning.
\end{IEEEkeywords}}

\maketitle

\IEEEdisplaynontitleabstractindextext

\IEEEpeerreviewmaketitle

\IEEEraisesectionheading{\section{Introduction}\label{sec:introduction}}

Training large deep neural networks requires massive amounts of
labeled training data. This fact hampers their application to domains
where training data is scarce and the process of collecting new
datasets is laborious and/or expensive. Recently, self-supervised
learning has received attention because it offers an alternative to
collecting labeled datasets. Self-supervised learning is based on the
idea of using an auxiliary task (different, but related to the
original supervised task) for which data is freely available and no
annotation is required. As a consequence, self-supervised learning can
be much more scalable. In~\cite{doersch2015unsupervised} the
self-supervised task is estimating the relative location of patches in
images. Training on this task allows the network to learn features
discriminative for semantic concepts. Other self-supervised tasks
include generating color images from gray scale images and vice
versa~\cite{larsson2017colorization,zhang2016colorful}, recovering a
whole patch from the surrounding pixels by
inpainting~\cite{pathak2016context}, and learning from equivalence
relations~\cite{noroozi2017representation}.

\begin{figure*}
  \centering
  \includegraphics[width=0.75\textwidth]{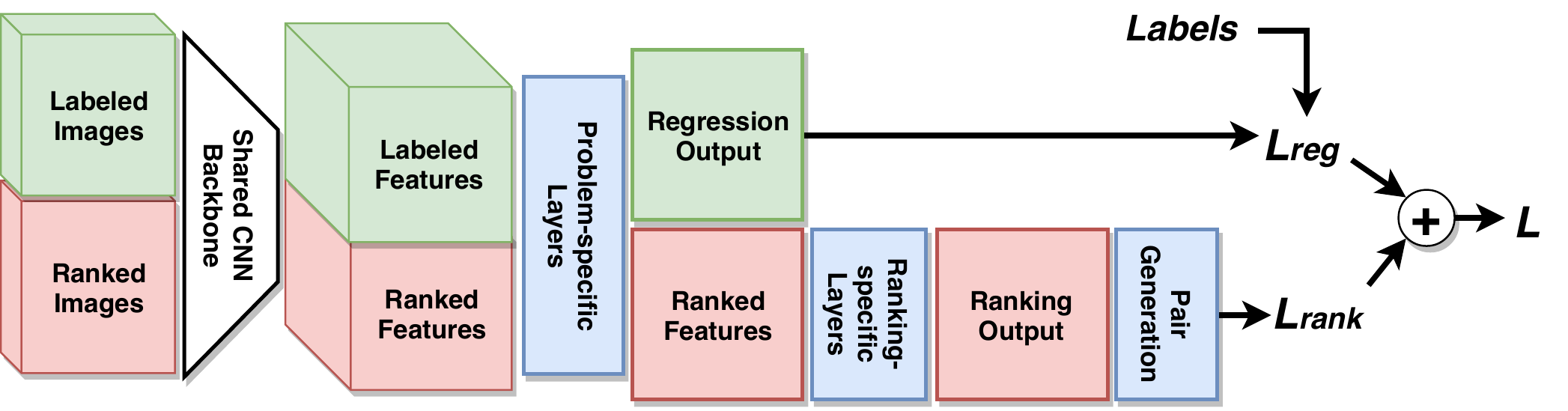}
\setlength{\tabcolsep}{5pt}
\caption{Self-supervised learning to rank. Our architecture is based
  on a shared CNN backbone (in white) to which we add problem-specific
  layers (in blue) that end in an output that solves the primary
  regression task, and layers (also in blue) that end in an output
  that solves a self-supervised ranking task (see
  Section~\ref{sec:learning}). Self supervision is provided by a pair
  generation module that is able to generate pairs with known relative
  ranks.}
\label{fig:ranking_examples}
\end{figure*}

In this paper, we investigate the use of ranking as a self-supervised
auxiliary task. In particular we consider regression problems in
computer vision for which it is easy to obtain ranked data
automatically from unlabeled data. By \emph{ranked data} we mean that
we know that for some samples the regression output is known to be
larger (or smaller) than for some others. In these cases the unlabeled
data, converted into ranked subsets of data, can be added in a
multi-task sense during training by minimizing an additional ranking
loss. The main advantage of our approach is that it allows adding
large amounts of unlabeled data to the training dataset, and as a
results train better deep neural networks. In addition, we show that
the ranked subsets of images can be exploited to performing active
learning by identifying which images, when labeled, will result in the
largest improvement of performance of the learning algorithm (here a
neural network). We consider two specific computer vision regression
problems to demonstrate the advantages of learning from ranked data:
Image Quality Assessment (IQA) and crowd counting. In
Fig.~\ref{fig:ranking_examples} we give an overview of the general
approach to use labeled and unlabeled (but ranked) data in a
multi-task network.

The first regression problem we consider is No-Reference Image Quality
Assesment (NR-IQA), where the task is to predict the perceptual
quality of images without using the undistorted image (also called the
reference image). This research field has also seen large
improvements in recent years due to the advent of Convolutional Neural
Networks (CNNs)~\cite{liang2016image,kang2014convolutional,kang2015simultaneous}.
The main problems these papers had to address is the lack of large
datasets for IQA. However, the annotation process for IQA image
datasets requires multiple human annotations for every image, and thus
the collection process is extremely labor-intensive and costly. As a
result, most available IQA datasets are too small to be effective for
training CNNs. We show how to automatically generate rankings for
the task of IQA, and how these rankings can be used to improve the
training of CNNs for NR-IQA. 

The second regression problem we consider is crowd counting. Crowd
counting is a daunting problem because of perspective distortion,
clutter, occlusion, non-uniform distribution of people, complex
illumination, scale variation, and a host of other scene-incidental
imaging conditions. Techniques for crowd counting have also seen
improvement recently due to the use of CNNs. These recent approaches
include scale-aware regression models~\cite{onoro2016towards},
multi-column CNNs~\cite{zhang2016single}, and switching
networks~\cite{Sam_2017_CVPR}. As with most CNN architectures,
however, these person counting and crowd density estimation techniques
are highly data-driven. Even modestly deep architectures for visual
recognition require massive amounts of labeled training data for
learning. For person counting, the labeling burden is even more
onerous than usual. Training data for person counting requires that
each individual person be meticulously labeled in training images. It
is for this reason that person counting and crowd density estimation
datasets tend to have only a few hundred images available for
training. As a consequence, the ability to train these sophisticated
CNN-based models suffers. We show how ranked sets of images can be
generated for the task of crowd counting, and how these can be
exploited to train deep networks.

In this work we study the use of ranking as a self-supervised
proxy task to leverage unlabeled data and improve the training of deep
networks for regression problems. The contributions of the paper are:
\begin{itemize}
\item We show that ranking tasks can be used as self-supervised proxy
  tasks, and how this can be exploited to leverage unlabeled data for
  applications suffering from a shortage of labeled data.
\item We propose a method for fast Siamese backpropagation which
  avoids the redundant computation common to training multi-branch
  Siamese network architectures. Similar observations have been made
  others~\cite{song2015deep,sohn2016improved} concurrently with our
  original work~\cite{liu2016master,liu2017rankiqa}.
\item We show that the ranking task on unlabeled data can be exploited
  as an active learning strategy to determine which images
  should be labeled to improve the performance of the network.
\item We demonstrate the advantages of the above contributes on two
  applications: Image Quality Assessment (IQA) and crowd counting. On both
  tasks we show how to effectively leverage unlabeled data and that
  this significantly improves performance over the state-of-the-art.
\end{itemize}

This paper is an extension of our previous work on
  No-reference IQA~\cite{liu2017rankiqa} and crowd
  counting~\cite{xialei2018crowd}. Beyond these previous works, here
  we propose a general framework for training from self-supervised
  rankings. In addition, we apply the multi-task approach
  from~\cite{xialei2018crowd} to the IQA problem, improving the
  results of our previous IQA method. We also include results on two
  new datasets for crowd counting, UCSD and WorldExpo'10. Finally, we
  show that self-supervised ranking tasks can also be used as a
  criterion for active learning. In this case the aim is to select
  which images to label from a large pool of unlabeled data. In
  experiments we show that this can significantly reduce the required
  labeling effort.

This paper is organized as follows. In the next section we review work
from the literature related to our work. In
section~\ref{sec:learning} we describe our general learning to rank
framework for self-supervised learning. In section~\ref{sec:iqa-loss} we
describe how to automatically generate rankings for image quality assessment. In section~\ref{sec:counting-loss} we show how the proposed framework can be applied to the problem of crowd counting. In sections~\ref{sec:iqa-experiments}~and~\ref{sec:counting-experiments} we give extensive
experimental evaluations for IQA and crowd counting, respectively. We
conclude in section~\ref{sec:conclusion} with a discussion of our
contributions and some indications of potential future research lines.

\section{Related work}
\label{sec:related}

In this section we review work from the literature on learning from
rankings, active learning, image quality assessment, and crowd counting.

\subsection{Learning from rankings}
Several works have studied how to learn \emph{to rank}, and they focus
on learning a ranking function from ground-truth
rankings~\cite{chen2009ranking,sculley2009large}.  They learn a ranking function from
ground-truth rankings by minimizing a ranking
loss~\cite{chen2009ranking}. This function can then be applied to rank
test objects. The authors of~\cite{sculley2009large} adapt the
Stochastic Gradient Descent method to perform pairwise learning to
rank. This has been successfully applied to large datasets. However, these approaches are very different from ours in which we aim to learn \emph{from}
rankings.

In a recent paper~\cite{noroozi2017representation} a method is
proposed where the self-supervised task is to learn to count. The
authors propose two proxy tasks -- scaling and tiling -- which guide
self-supervised training. The network learn to count visual primitives
in image regions. It is self-supervised by the fact that the number of
visual primitives is not expected to change under scaling, and that
the sum of all visual primitives in individual tiles should equal the
total number of visual primitives in the whole image. Unlike our
approach, they do not consider rankings of regions and their counts
are typically very low (several image primitives). Also, their final
tasks do not involve counting but rather unsupervised learning of
features for object recognition. 

\subsection{Active learning}
Active learning~\cite{settles2012active} is a machine learning procedure that reduces the cost of annotation by actively selecting
the best samples to label among the abundantly available unlabeled
data.  Active learning is well-motivated in many modern machine
learning problems where data may be abundant, but labels are scarce or
expensive to obtain. Since massive amounts of data are required to
train deep neural networks, informative samples are more valuable to
annotate instead of annotating randomly picked samples. Active
learning has been explored in many applications such as image
classification~\cite{zhou2017fine}, object
detection~\cite{papadopoulos2016we} and image
segmentation~\cite{yang2017suggestive}.

There are several ways to approach the active learning
problem. Uncertainty sampling~\cite{lewis1994heterogeneous} is the
simplest and most commonly used that samples the instance whose
prediction is least confident. Margin
sampling~\cite{scheffer2001active} aims to correct for the
shortcomings of uncertainty sampling by examining the difference
between second and first most likely labels for candidate samples. A
more general strategy considers all prediction probabilities using
entropy. Expected Model Change~\cite{settles2008multiple} is a metric
that measures change in gradient by calculating the length as an
expectation over the possible labelings. However, these approaches
only consider the uncertainty of instances and ignore the
representatives of the underlying distribution. 
The method proposed in~\cite{settles2008active} addresses this issue by computing the similarity between the candidate instance and all other samples in the training set. Note that these active learning approaches are more about classification problems, while we are primarily interested in
regression.

In this work we show how to apply active learning to image quality
assessment and crowd counting -- two tasks for which image annotation
is expensive. Instead of using the above approaches, we measure the
informativeness of unlabeled instances via mistakes made by the
network on a self-supervised ranking proxy task. Performance on this proxy task is guaranteed to be consistent with the main task, and our hypothesis is that the instances with more mistakes on the proxy task vary more from
the training set. Training on such instances should allows us to
increase the generalizing capacity of neural networks.

\subsection{Image Quality Assessment}
We briefly review the IQA literature related to our approach. We focus
on recent deep learning based methods for distortion-generic, No-reference IQA since it is more generally applicable than the other IQA research lines. 

In recent years several works have used deep learning for
NR-IQA~\cite{bianco2016use,kang2014convolutional,kang2015simultaneous}.
One of the main drawbacks of deep networks is the need for large
labeled datasets, which are currently not available for NR-IQA
research. To address this problem Kang et
al.~\cite{kang2014convolutional} consider small $32 \times 32$ patches
rather than images, thereby greatly augmenting the number of training
examples. The authors of~\cite{bosse2016deep,kang2015simultaneous}
follow the same pipeline. In~\cite{kang2015simultaneous} the authors
design a multi-task CNN to learn the type of distortions and image
quality simultaneously. Bianco at al.~\cite{bianco2016use} propose to
use a pre-trained network to mitigate the lack of training data. They
extract features from a pre-trained model fine-tuned on an IQA
dataset. These features are then used to train an SVR model to map
features to IQA scores.

There are other works which, like us, apply rankings in the
  context of NR-IQA.  Gao et al. ~\cite{gao2015learning} generate
  pairs in the dataset itself by using the ground truth scores with a
  threshold and combine different hand-crafted features to represent
  image pairs from the IQA dataset. Xu et al.~\cite{xu2017multi}
  propose training a specific model for each distortion type in a
  multi-task learning model. Instead of using the ground truth in the
  dataset, they learn a ranking function. The most relevant work is
  from Ma et al. ~\cite{ma2017dipiq}, which was published concurrently
  with our work on NR-IQA~\cite{liu2017rankiqa}.  Like us, they use
  learning-to-rank to deal with the extremely limited ground truth
  data for training. However, they generate pairing data by using
  other Full-reference IQA methods with a threshold like
  in~\cite{gao2015learning}, while our approach does not require other
  methods and operates in a self-supervised manner. Another difference
  is that we use a knowledge transfer technique for further
  fine-tuning on the target dataset with the same network, which can
  be also learned in a multi-task, end-to-end way, while they train an
  independent regression model based on the feature representations to
  perform the prediction. In addition to these differences, we show
  results on a larger number of distortion types instead of only
  working on the four main distortions~\cite{ma2017dipiq}.

In our paper, we propose a radically different approach to address the
lack of training data: we use a large number of automatically
generated rankings of image quality to train a deep network. This
allows us to train much deeper and wider networks than other methods
in NR-IQA which train directly on absolute IQA data.

\subsection{Crowd counting}

We focus on deep learning methods for crowd counting in still images. For a more complete review of the literature on crowd counting, including classical approaches, we refer the reader to~\cite{sindagi2017survey}.

As introduced in the review of~\cite{sindagi2017survey}, CNN-based
approaches can be classified into different categories based on the
properties of the CNN. Basic CNNs
incorporate only basic CNN layers in their networks. The approaches
in~\cite{fu2015fast,wang2015deep} use the AlexNet
network~\cite{krizhevsky2012imagenet} to map from crowd scene patches
to global number of people by changing the output of AlexNet from 1000
to 1. The resulting network can be trained end-to-end.
Due to the large variations of density in different images, recent
methods have focused on scale-awareness. The method proposed in
\cite{zhang2016single} trains a multi-column based architecture (MCNN)
to capture the different densities by using different sizes of kernels
in the network. Similarly, the authors of~\cite{onoro2016towards}
propose the Hydra-CNN architecture that takes different resolutions of
patches as inputs and has multiple output layers (heads) which are
combined in the end. Most recently, in~\cite{Sam_2017_CVPR} the
authors propose a switching CNN that can select an optimal head
instead of combining the information from all network heads. Finally,
context-aware models are networks that can learn from the context of
images. In~\cite{fu2015fast,sindagi2017generating} the authors propose
to classify images or patches into one of five classes: very high
density, high density, medium density, low density and very low
density. However, the definition of these five classes varies across
datasets and must be carefully chosen using knowledge of the
statistics of each dataset.

Although CNN-based methods have achieved great success in crowd
counting, due to lack of labeled data it is still challenging to train
deep CNNs without over-fitting. The authors of~\cite{zhang2015cross}
propose to learn density map and global counting in an alternating
sequence to obtain better local optima. The method 
in~\cite{kang2016crowd} uses side information like ground-truth camera
angle and height to help the network to learn. However, this side
information is expensive to obtain and is not available in most
existing crowd counting datasets.

There is some interesting recent work on CNNs for crowd
  counting. CSRNet~\cite{li2018csrnet} consists of two components: a
  convolutional neural network as 2D feature extractor and a dilated
  CNN for estimating a density map to yield larger receptive fields
  and to replace pooling operations. DecideNet~\cite{liu2018decidenet}
  starts by estimating the crowd density using detection and
  regression separately. It then assesses the reliability of these two
  estimates with an attention module. Shen et al.~\cite{shen2018crowd}
  propose using a U-net structure to generate a high quality density
  map with an adversarial loss. Shi et al. ~\cite{shi2018crowd}
  formulate a single ConvNet as ensemble learning. Marsden et
  al.~\cite{marsde2018people} adapt object counting models to new
  visual domains like cell counting and penguins counting. Both
  works~\cite{ranjan2018iterative,cao2018scale} propose generating a
  high resolution density map. Ierees et
  al.~\cite{idrees2018composition} propose solving the problems of
  counting, density map estimation and localization
  simultaneously. Laradji et al. ~\cite{laradji2018blobs} propose a
  detection-based method that does not need to estimate the size and
  shape of the objects.

In our paper, we show how a large number of unlabeled crowd data can
improve the training of crowd counting networks. We automatically
generate rankings from the unlabeled images, which are used during the
training process in the self-supervised proxy task.

\section{Learning from rankings}

\label{sec:learning}

In this section we lay out a general framework for our approach, then
describe how we use a Siamese network architecture to learn from
rankings. In section~\ref{sec:efficient-siamese} we show how
backpropagation for training Siamese networks from ranked samples can
be made significantly more efficient. Finally, in section~\ref{sec:AL}
we show how the ranking proxy task can be used as an active learning
algorithm to identify which are the most important images to label
first.

\subsection{Ranking as a self-supervised proxy task}
Regression problems consist of finding a mapping function between
input variables and a continuous output variable. It is a vital area
of research in machine learning, and many important problems in
computer vision are regression problems. The mapping function is
typically learned from a training dataset of labelled data which
consists of pairs of input and output variables. The complexity of the
mapping function that can be learned is limited by the number of
labeled examples in the training set. In this article, we are
interested in using deep convolutional neural networks (CNNs) as
mapping functions, and images as input data.

For some regression problems it can be easy to obtain a \emph{ranked
  dataset}. Such a dataset contains relative information between pairs
of input examples, describing which of the two is larger. For image
quality assessment (IQA) it is easy to generate ranked images by
applying different levels of distortions to an image. As an example,
given a reference image we can apply various levels of Gaussian
blur. The set of images which is thus generated can be easily
\emph{ranked} because we do know that adding Gaussian blur (or any
other distortion) \emph{always} deteriorates the quality score. Note
that in such a set of ranked images we do not have any absolute IQA
scores for any images -- but we do know for any pair of images
\emph{which is of higher quality}. See Fig.~\ref{fig:IQA-datagen} (bottom) for
an illustration of this.

For the crowd counting problem, we can obtain ranked sets of images by
comparing parts of the same image which are contained within each
other: an image which is contained by another image will contain the
same number or fewer persons than the larger image. This fact can be
used to generate a large dataset of ranked images from unlabeled crowd
images. See Fig.~\ref{fig:counting-datagen} (bottom) for an illustration of this
process for crowd counting.

In the following sections we will show that ranked data can be used to
train networks for regression problems. We are especially interested
in domains where the ranked data can be automatically generated from
images of the problem domain without requiring any additional human
labeling. This allows us to create large dataset of ranked data. We
consider regression to be the \emph{principal task} of the network,
and we refer to the ranking task as a \emph{self-supervised proxy
  task}. It is self-supervised since the ranking task is an additional
task for which data is freely available and no annotation is required.

\subsection{Multi-task regression and ranking}
In this section, we formalize the problem of training from both
labeled and ranked data for regression problems. We consider a
regression problem where we have a dataset of observed data in pairs:
\begin{equation}
D = \left\{ {\left( {{\bf{x}}_1 ,y_1 } \right),\left( {{\bf{x}}_2 ,y_2 } \right)...,\left( {{\bf{x}}_n ,y_n } \right)} \right\},
\end{equation}
where $\mathbf{x}_i$ are images and
$y_i\in \mathbb{R}$. The input images $\mathbf{x}_i$ and target
variables $y_i$ are assumed to be related by some unknown function
$f(\mathbf{x}_i) = y_i$. The aim is to find a function
$\hat{f}(\mathbf{x}; \theta)$ with parameters $\theta$ that captures
(and generalizes) the relationship between input $\mathbf{x}$ and
output $y$. The parameters $\theta$ of the regression function
$\hat{f}$ are usually fit by minimizing an empirical risk over
training examples $D$, for example the squared Euclidean loss:
\begin{equation}
L_{reg} = \sum_{i=1}^n ( \hat{f}( \mathbf{x}_i ;\theta ) - y_i )^2.
\end{equation}
In our formulation, $\hat{f}$ is a deep Convolutional Neural Network (CNN),
and minimizing $L_{reg}$ is done with stochastic gradient descent
(SGD). We refer to the regression task as the \emph{principal task},
since the final objective is to accurately estimate such a regression.

We also assume that we have a function $h(\cdot, \varphi)$ which we
can apply to input images. These functions are special in that the
parameter space is ordered and that for any parameters
$\varphi_i, \varphi_j$ and any image $\mathbf{x}$:
\begin{eqnarray}
  \varphi_i \leq \varphi_j \Rightarrow f(h(\mathbf{x}; \varphi_{i})) \leq f(h(\mathbf{x}; \varphi_{j})).
  \label{eq:ordering}
\end{eqnarray}
What this means is that we have a \emph{partial order} in the
parameter space that induces an ordering in the proxy task space.  An
example of a possible function $h$ would be adding an image distortion
parameterized by a single scalar number: \emph{increasing} this
parameter in $h$ implies \emph{decreasing} image quality in
$h(\mathbf{x}, \varphi)$. For crowd counting we can parameterize $h$
using rectangular crops: if a rectangle $\varphi_i$ is \emph{entirely
  contained} in rectangle $\varphi_j$, then the number of persons in
$h(\mathbf{x}; \varphi_{i})$ is less than or equal to
$h(\mathbf{x}; \varphi_{j})$. In other words:
$f(h(\mathbf{x}; \varphi_{i})) \leq f(h(\mathbf{x};
\varphi_{j}))$.

We can now apply this function $h$ to generate an auxiliary dataset
$E$ consisting of \emph{ranked images}. Importantly, the ordering of
the parameter space and the application of $h$ is independent of any
labeling of training data. The dataset $E$ could contain images
${\bf x}_i$ which are present in dataset $D$ but in addition it could
also contain images not in $D$ and for which we have no
annotations. We can use any input data $\mathbf{x}_i$ relevant to the
domain in order to generate the dataset $E$ of ranked images. Since
$f$ respects this ranking condition under application of functions
$h$, we can use the $h$ functions to generate training data for
learning $\hat{f}$.

From dataset $E$ we can train a network by minimizing the ranking
hinge loss according to:
\begin{equation}
  \label{eq:ranking}
  L_{rank} = \hspace{-0.1in} \sum_{\substack{\mathbf{z} \in E \\ \varphi_i \leq \varphi_j}} \hspace{-0.08in} \max (0, \hat{f}(h(\mathbf{z}; \varphi_i); \theta) - \hat{f}(h(\mathbf{z}; \varphi_j); \theta) + \varepsilon)
\end{equation}
where $\varepsilon$ is a margin, and $\mathbf{z} \in E$ are
(potentially unlabeled) images transformed into \emph{ranked} images
after applying $h(\cdot, \varphi_{i})$ and $h(\cdot, \varphi_{j})$
for $\varphi_i \leq \varphi_j$. Training a network with
Eq.~(\ref{eq:ranking}) yields a network which can \emph{rank} images,
and we will refer to the task of ranking as the \emph{self-supervised
  proxy task}.

The most common approach to minimizing losses like $L_{rank}$ uses a
Siamese network~\cite{chopra2005learning}, which is a network with two
identical branches connected to a loss module. The two branches
share weights during training. Pairs of images and labels are the
input of the network, yielding two outputs which are passed to the
loss. The gradients of the loss function with respect to all model
parameters are computed by backpropagation and updated by the
stochastic gradient method (e.g. SGD). For problems where both labeled
data (like in dataset $D$) and ranked data (like in dataset $E$) are
present, we can optimize the network using \emph{both} sources of data
using a \emph{multi-task loss}:
\begin{equation}
L = L_{reg}  + \lambda L_{rank} \label{eq:multitask_loss}
\end{equation}
where $\lambda$ is a tradeoff parameter that balances the relative
weight of the losses. It is important to note here that we consider
the same function $\hat{f}( \cdot ;\theta )$ for the regression and
the ranking loss. In practice this means that the three networks, one
for regression, and the two networks used in the Siamese network, have
the same architecture and share their parameters.

One could also consider different ways to combine both the labeled
dataset and the ranking dataset. In earlier work we investigated using
the ranking dataset to train initial weights of the network, after
which we used the labeled data for
fine-tuning~\cite{liu2017rankiqa}. This is the approach which is used
by almost all self-supervised methods in computer
vision~\cite{doersch2015unsupervised,pathak2016context,zhang2016colorful,noroozi2017representation,liu2017rankiqa}.
However, in~\cite{xialei2018crowd} we found this to be inferior to
using the multi-task loss, and in this work only consider
multi-task formulations like in~Eq.~(\ref{eq:multitask_loss}).

\subsection{Efficient Siamese backpropagation}
\label{sec:efficient-siamese}

One drawback of Siamese networks is redundant computation. Consider
all possible image pairs constructed from three images. In a standard
implementation all three images are passed twice through the network,
because they each appear in two pairs. Since both branches of the
Siamese network are identical, we are essentially doing twice the work
necessary since any image need only be passed \emph{once} through the
network. It is exactly this idea that we exploit to render
backpropagation more efficient for Siamese network training. In fact,
nothing prevents us from considering \emph{all} possible pairs in a
mini-batch, with hardly any additional computation. We add a new layer
to the network that generates all possible pairs in a mini-batch at
the end of the network right before computing the loss. This
eliminates the problem of pair selection and boosts efficiency. At the
end of this section we discuss how our approach compares to
works~\cite{song2015deep,sohn2016improved} published concurrently with
ours and observed similar efficiency gains.

To appreciate the speed-up of efficient Siamese backpropagation
consider the following. If we have one reference image distorted from
which we have generated $n$ ranked images using $h$, then for a
traditional implementation of the Siamese network we would have to
pass a total of $n^2-n$ images through the network -- which is twice
the number of pairs you can generate with $n$ images. Instead we
propose to pass all images only \emph{once} and consider all possible
pairs only in the loss computation layer.  This reduces computation to
just $n$ images passed through the network. Therefore, in this case
the speed-up is equal to: $\frac{n^2-n}{n} = n-1$. In the best
scenario $n$ is equal to the number of images in the
mini-batch, and hence the speed-up of this method would be in the
order of the mini-batch size. Due to the high correlation among the
set of all pairs in a mini-batch, we expect the final speedup in
convergence to be lower.

To simplify notation in the following, assume we have an image
$\mathbf{z}$ and two transformation parameters
$\varphi_i \le \varphi_j$. Letting
$\hat{y}_i = \hat{f}(h(\mathbf{z}, \varphi_i); \theta)$, the
contribution to the ranking loss $L_{rank}$ of these two images can be
written as:
\begin{equation} \label{eq:general}
g(\hat{y}_i,\hat{y}_j) = \max (0, \hat{y}_i - \hat{y}_j + \varepsilon),
\end{equation}
The gradient of this term from $L_{rank}$ with respect to the model
parameters $\theta$ is:
\begin{equation}
  \nabla_{\theta} g=
  \frac{ \partial g(\hat{y}_i,\hat{y}_j) }
  {\partial \hat{y}_i }
   \nabla_{\theta} \hat{y}_i
  +
  \frac{ \partial g(\hat{y}_i,\hat{y}_j) }
  {\partial \hat{y}_j }
   \nabla_{\theta} \hat{y}_j.
\end{equation}
This gradient of $g$ above is a sum since the model parameters are
shared between both branches of the Siamese network and $\hat{y}_i$
and $\hat{y}_j$ are computed using exactly the same
parameters.

Considering all pairs in a mini-batch of size $M$, the loss $L_{rank}$
from Eq.~(\ref{eq:ranking}) can then be written as:
\begin{equation}
L_{rank} = \sum_{i=1}^{M}{ \sum _{ j > i }^{ M }{ g(\hat{y}_i,\hat{y}_j) }}.
\end{equation}
The gradient of the mini-batch loss with respect to parameter $\theta$
can then be written as:
\begin{equation}
\nabla_{\theta} L_{rank}  =
\sum_{ i=1 }^{ M } \sum_{ j > i }^M
 \frac{ \partial g(\hat{y}_i,\hat{y}_j) }
      { \partial \hat{y}_i}
      \nabla_{\theta} \hat{y}_i
 +
 \frac{ \partial g(\hat{y}_i,\hat{y}_j) }
      { \partial \hat{y}_j }
      \nabla_{\theta} \hat{y}_j.
\label{eq:loss2}
\end{equation}
We can now express the gradient of the loss function of the mini-batch
in matrix form as:
\begin{equation}
\nabla_\theta  L_{rank} = \left[ {\begin{array}{*{20}c}
   {\nabla _\theta  \hat y_1 } & {\nabla _\theta  \hat y_2 } &  \ldots  & {\nabla _\theta  \hat y_M }  \\
\end{array}} \right]P{\bf{1}}_M \label{eqn:minibatch-gradient},
\end{equation}
where $\mathbf{1}_M$ is the vector of all ones of length $M$. For a
standard single-branch network, we would average the gradients for all
batch samples to obtain the gradient of the mini-batch. This is
equivalent to setting $P$ to the identity matrix
in Eq.~(\ref{eqn:minibatch-gradient}) above. For Siamese networks where we
consider all pairs in the mini-batch we obtain Eq.~(\ref{eq:loss2}) by
setting $P$ to:
\begin{equation}
\arraycolsep=1.5pt
P = \left[
    \begin{matrix}
      0 & \frac { \partial g(\hat{y}_1,\hat{y}_2) }{ \partial \hat{y}_1 }  & \cdots  & \frac{\partial g(\hat{y}_1,\hat{y}_M)}{ \partial \hat{y}_1 }  \\
      \frac { \partial g(\hat{y}_1,\hat{y}_2) }{ \partial \hat{y}_2 } & 0 & \cdots & \frac { \partial g(\hat{y}_2,\hat{y}_M) }{ \partial \hat{y}_2 } \\
      \vdots  & \vdots  & \ddots  & \vdots  \\
      \frac { \partial g(\hat{y}_1,\hat{y}_M) }{ \partial \hat{y}_M } & \cdots &   \cdots & 0 \end{matrix} \right] \label{eq:P_matrix}.
\end{equation}
For the ranking hinge loss we can write:
\begin{equation}\label{eq:fast}
  P =  \left[ \begin{matrix} 0 & { a }_{ 12 } & \cdots  & { a }_{ 1M } \\ { a }_{ 21 } & 0 & \cdots  & { a }_{ 2M } \\ \vdots  & \vdots  & \ddots  & \vdots  \\ { a }_{ M1 } & \cdots  & { a }_{ M(M-1) } & 0 \end{matrix} \right],
\end{equation}
where
\begin{eqnarray}
\label{eqn:coefficients}
 a_{ij} &=&
\begin{cases}
0 & \text{if } l_{ij} \left( \hat{y}_i -  \hat{y}_j \right) + \varepsilon  \le  0   \\
l_{ij} & \text{otherwise}
\end{cases}
\end{eqnarray}
and $l_{ij}$ is used to indicate the ordering of the $\varphi$
parameters used to generate the $M$ images to which $\hat{f}$ was applied to
derive outputs $\hat{y}_i$ and $\hat{y}_j$:
\begin{eqnarray}
l_{ij} &=&
\begin{cases}
\ \ \ 1 & \mbox{if } \varphi_i \leq \varphi_j \\
-1 & \mbox{if } \varphi_i > \varphi_j \\
\ \ \ 0 & \mbox{if $\varphi_i$ and $\varphi_j$ are not comparable.}
\end{cases}
\end{eqnarray}
The above analysis works for different parameter settings $\varphi$ on the same source image. When  considering multiple source images in a mini-batch, only different parameter settings $\varphi$ on the same image are considered comparable when defining $l_{ij}$.

Generally, the complexity of training Siamese networks is ameliorated
via different pair sampling
techniques. In~\cite{simo2015discriminative}, the authors propose a
hard positive and hard negative mining strategy to forward propagate a
set of pairs and sample the highest-loss pairs for
backpropagation. However, hard mining comes with a high computational
cost (they report an increase of up to 80\% of total computation
cost). In~\cite{schroff2015facenet} the authors propose semi-hard pair
selection, arguing that selecting hardest pairs can lead to bad local
minima. In~\cite{wang2015unsupervised} the authors take a batch of
pairs as input and choose the four hardest negative samples within the minibatch. In parallel with our work on fast backpropagation for Siamese
networks~\cite{liu2016master,liu2017rankiqa} several similar methods
have been developed~\cite{song2015deep,sohn2016improved}. To solve for
bad local optima, \cite{song2015deep} optimize a smooth upper bound
loss function. This is implemented by considering all possible pairs
in a mini-batch after forwarding the images through the network. In
\cite{sohn2016improved}, the
$N$-pair Loss is proposed to compute pairwise similarity within the
batch to construct
$N-1$ negative examples instead of one in triplet loss. Hard negative
class mining is applied to improve convergence speed.  Both these
works, like ours, prevent the redundant computation which is
introduced by the multiple branches in the Siamese network.

\subsection{Active learning from rankings}
\label{sec:AL}

The objective of active learning is to reduce the cost of labeling by
prioritizing the most informative samples for labeling first. Rather
than labeling randomly selected examples from a pool of unlabeled
data, active learning methods analyze unlabeled data with the goal of
identifying images considered difficult and that therefore are more
valuable if labeled. Especially for deep networks, which require many
examples, as well as for applications for which labeling is very
costly, active learning is an important and active area of research.

We show here how the self-supervised proxy task can be leveraged for
active learning. For this purpose we define a function
$C(\mathbf{x}_i)$ which estimates the certainty of the current network
on a specific image $\mathbf{x}_i$. This estimate allows us to order
the available unlabeled dataset according to certainty. Labeling the
images for which the algorithm is \emph{uncertain} is then expected to
yield larger improvement in performance of the network than just
randomly adding images.

The certainty function $C$ is defined as:
\begin{equation}
  C(\mathbf{z}; \theta) = \frac{1}{K} \hspace{-0.07in} \sum_{\varphi_i \leq \varphi_j} \hspace{-0.07in}
  T( \hat{f} ( h ( \mathbf{z}; \varphi_i); \theta ) < \hat{f} ( h(\mathbf{z} ;\varphi_j); \theta ) ) 
\label{eq:ranked}
\end{equation}
where $T(s) = 1$ if predicate $s$ is true or false otherwise, and $K$ is
the number of sampled parameter pairs $(\varphi_i, \varphi_j)$ applied
to each image $\mathbf{z}$. As described in the previous section, the
pairs of parameters $\varphi_i \leq \varphi_j$ can be used to generate
ranked images via the function $h$. Again, these pairs can be
automatically computed and no annotation is required.

By design, we know that $0 \le C ( \mathbf{x}_i ) \le 1$. Given an
unlabeled dataset and a current state of the trained network
$\hat{f}(\cdot,\hat{\theta})$, we perform the proxy task for a total
of $K$ times on each image and then compute $C$. We then label images
starting from low confidence to high confidence. The process is
detailed in Algorithm~\ref{table:algorithm}.

\begin{algorithm}[tb]
\caption{: Active learning loop. }
\label{table:algorithm}
\begin{algorithmic}
  \STATE\textbf{Input:}
  \vspace{-0.08in}
  {\setlength{\mathindent}{0cm}
  \begin{equation*}
    \begin{array}{ll}
      D & = \set{ {\left( {{\bf{x}}_1 ,y_1 } \right), \ldots, \left( {{\bf{x}}_n ,y_n } \right)} }: \hspace{-0.0in} \mbox{ labeled samples} \\
      E & = \set{ \mathbf{z}_{1}, \ldots ,\mathbf{z}_{m} }: \hspace{-0.0in} \mbox{ unlabeled samples}
    \end{array}
  \end{equation*}}

\vspace{-0.1in}
\STATE\textbf{Require:}
\vspace{-0.08in}
{\setlength{\mathindent}{0cm}
\begin{equation*}
  \begin{array}{lll}
    \theta_0 \hspace{-0.1in} &:&  \hspace{-0.1in} \mbox{ initial network parameters} \\
    T \hspace{-0.1in} &:& \hspace{-0.1in} \mbox{ number of active learning cycles} \\
    S \hspace{-0.1in} &:& \hspace{-0.1in} \mbox{ number of images added in each cycle}
  \end{array}
\end{equation*}}
\vspace{-0.1in}

\textbf{for} $t = 1:T$ \\
\vspace{-0.2in}
{
\setlength{\mathindent}{0.2in}
\begin{equation*}
  \arraycolsep=1.4pt
  \begin{array}{llll}
    \theta_t & \leftarrow & \mathbf{train}(\mathcal{D}, \theta_{t-1}) & \mbox{ \footnotesize  // Train network on $D$.}\\
    D^S & \leftarrow & \mathbf{label}(E, S, \theta_t) & \mbox{ \footnotesize // Evaluate Eq.~(\ref{eq:ranked}) for all samples in E,}\\
        &            &                                & \mbox{ \footnotesize // and Label $S$ least confident samples.} \\
    D   & \leftarrow & D \bigcup D^S & \mbox{ \footnotesize // Update labeled set.} \\
    E   & \leftarrow & E \setminus D^S & \mbox{ \footnotesize // Update unlabeled set.}
  \end{array}
\end{equation*}
}
\vspace{-0.1in}

\textbf{end for}

\end{algorithmic}
\end{algorithm}

\section{Image Quality Assessment by Learning to Rank}
\label{sec:iqa-loss}
In this section we apply the framework proposed in the previous
section to the problem of Image Quality Assessment
(IQA)~\cite{wang2002image}. IQA aims to automatically predict the
perceptual quality of images. IQA estimates should be highly
correlated with quality assessments made by a range of very many human
evaluators (commonly referred to as the Mean Opinion Score
(MOS)~\cite{sheikh2006statistical,ponomarenko2013color}).  We focus
here on no-reference IQA (NR-IQA), which refers to the case where the
undistorted image (called reference image) is not available during the
quality assessment.

\begin{figure}
  \begin{tabular}{c}
  \includegraphics[width=\columnwidth]{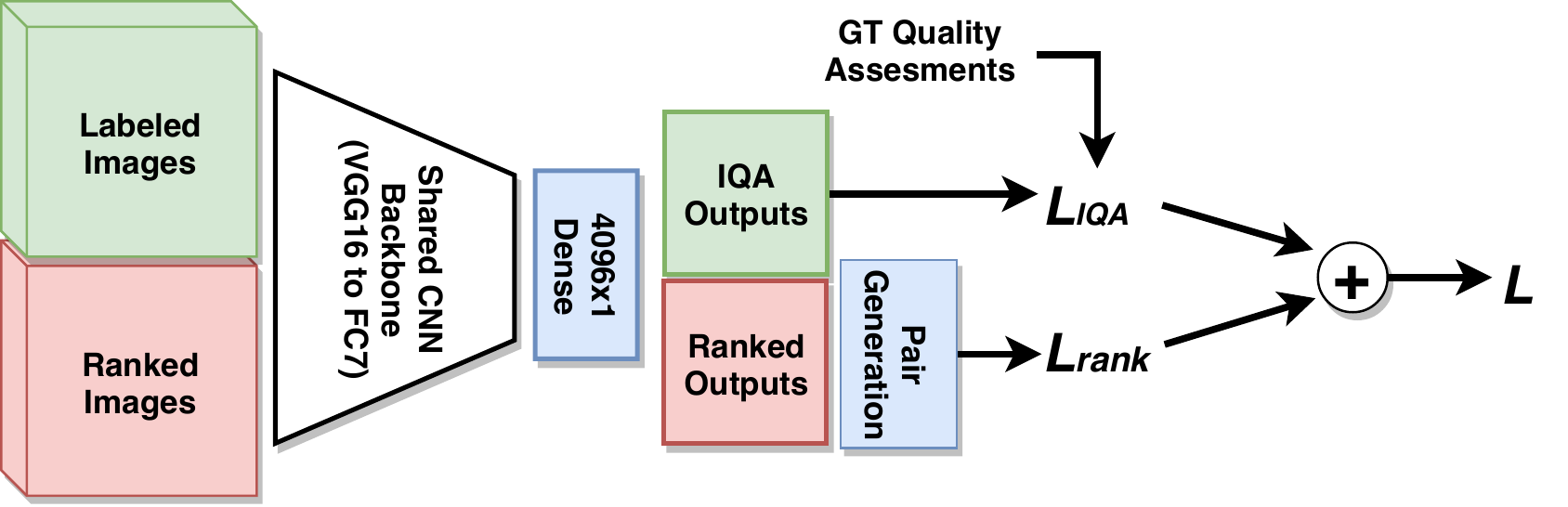} \\
  \includegraphics[width=\columnwidth]{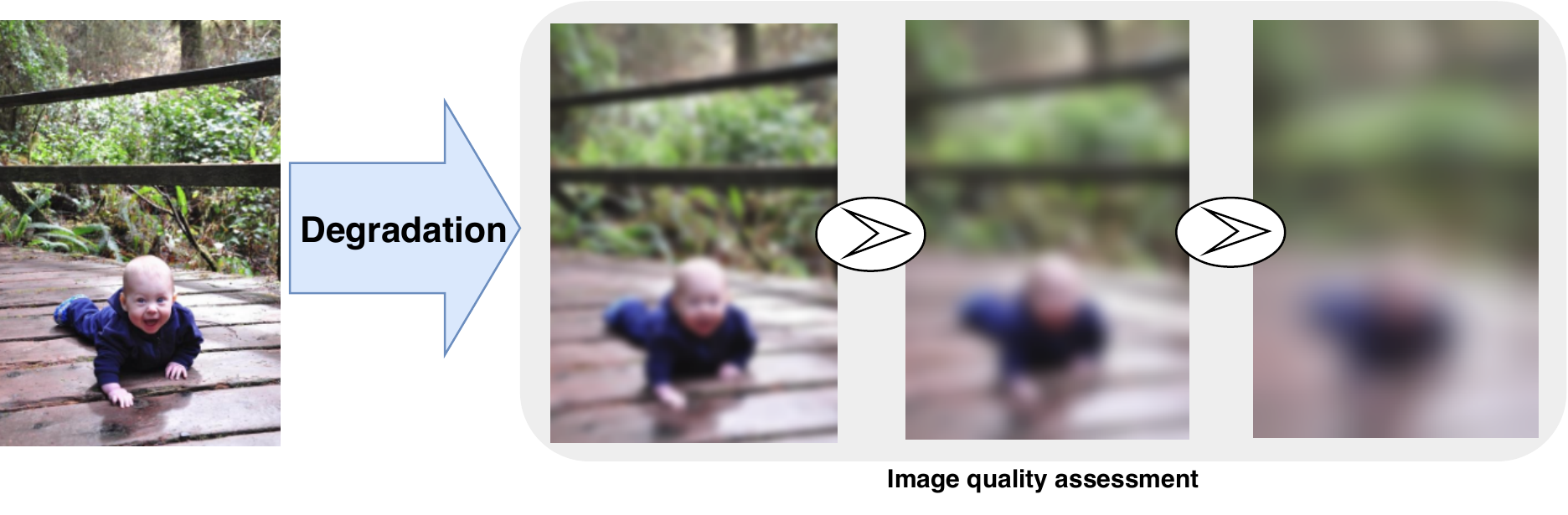}
 \end{tabular}
 \caption{
   Network architecture and ranked pair generation for
   IQA. \textbf{Top}: our network for no-reference IQA uses a VGG16 network
   pretrained on ImageNet. We decapitate the network and replace the
   original head with a new fully-connected layer generating a single
   output. \textbf{Bottom}: pairs with known ranking are generated by
   distorting images with standard, parametric distortions. Increasing
   the distortion level guarantees that the images are of
   progressively worse quality.
 }
 \label{fig:IQA-datagen}
\end{figure}

The application of CNNs to IQA has resulted in significant
improvements compared to previous hand-crafted
approaches~\cite{liang2016image,kang2014convolutional,kang2015simultaneous}. 
These methods had to train their networks on the small available datasets for IQA. Further improvements would be expected if larger datasets were made available. However, the annotation of IQA images is a labor intensive task, which requires multiple human annotators for every image. It is therefore an interesting application field to
evaluate our framework, since by adding ranking as a proxy task, we
are able to add unlabeled data during the training process.

We first discuss existing datasets for IQA and how to automatically
generate IQA rankings. Then we introduce the network which we train
for the IQA task, together with some application-specific training
choices.

\subsection{IQA datasets}
We perform experiments on two standard IQA datasets:
\begin{itemize}[leftmargin=*]
\setlength\itemsep{0em}
\item \textbf{LIVE}~\cite{live2}: Consists of 808 images generated
  from 29 original images by distorting them with five types of
  distortion: Gaussian blur (GB), Gaussian noise (GN), JPEG
  compression (JPEG), JPEG2000 compression (JP2K) and fast fading
  (FF). The ground-truth Mean Opinion Score for each image is in the
  range [0, 100] and is estimated using annotations by 161 human
  annotators.

\item \textbf{TID2013}~\cite{ponomarenko2013color}: Consists of 25
  reference images with 3000 distorted images from 24 different
  distortion types at 5 degradation levels. Mean Opinion Scores are in
  the range [0, 9]. Distortion types include a range of noise,
  compression, and transmission artifacts. See the original
  publication for the list of specific distortion types.

\end{itemize}

\subsection{Generating ranked image sets for IQA}
The IQA datasets LIVE and TID2013 are derived from only 29 and 25
original images, respectively. Deep networks trained on so few
original images will almost certainly have difficulty generalizing to
other images. Here we show how to automatically generate rankings from
arbitrary images which can be added during the training process
(i.e. we show what function $h$ in Eq.~(\ref{eq:ranking}) we use to
generate IQA rankings).

Using an arbitrary set of images, we can synthetically generate
deformations of these over a range of distortion intensities.  As an
example, consider Fig.~\ref{fig:IQA-datagen} (bottom) where we have
distorted an image with increasing levels of Gaussian noise. Although
we do not know the absolute IQA score, we do know that images which
are \emph{more} distorted should have a \emph{lower} score than images
which are less distorted. This fact can be used to generate huge
datasets of ranked images. We can use a large variety of images and
distortions to construct the datasets of ranked images.

We use the Waterloo~\cite{magroup} dataset, which consists of 4,744
high quality natural images carefully chosen from the Internet, to
generate a large ranking dataset. To test on the LIVE database, we
generate four types of distortions which are widely used and common:
Gaussian Blur (GB), Gaussian Noise (GN), JPEG, and JP2K. We apply
these distortions at five levels, resulting in a total of 20
distortions for all images in the Waterloo dataset. To test on
TID2013, we generate 17 out of a total of 24 distortions at five
levels, yielding a total of 85 distortions per image (see the Appendix
for more
details). 
There were seven distortions which we could not generate, however we
found that adding the other distortions also resulted into
improvements for the distortions for which we could not generate
rankings.

\subsection{IQA network}
For the IQA problem we have a training dataset with images
$\mathbf{x}_i$ and ground truth image quality $y_i$. After generating
the ranked image dataset we can train an IQA network. As a CNN
backbone we use the VGG-16 network, only changing the last layer to
output a single IQA score. With respect to the basic architecture
given in Fig.~\ref{fig:ranking_examples}, here we specify the
architecture we use for IQA estimation as shown in
Fig.~\ref{fig:IQA-datagen} (top).  For the labeled data we use the
Euclidean distance between the prediction of the network $\hat{y}_i$
and the ground truth as the regression loss:
\begin{equation}
L_{IQA}(y_i, \hat{y}_i) = \frac { 1 }{N } \sum _{ i=1 }^{ N } (y_{i}-\hat{y}_{i})^2,
\end{equation}
and will optimize this loss jointly with the ranking loss $L_{rank}$
as proposed in Eq.~(\ref{eq:multitask_loss}).

We randomly sample sub-images from the original high resolution
images. We do this instead of scaling to avoid introducing distortions
caused by interpolation or filtering. The size of sampled images is
determined by each network. However, the large size of the input
images is important since input sub-images should be at least 1/3 of
the original images in order to capture context information. This is a
serious limitation of the patch sampling
approach~\cite{kang2014convolutional,kang2015simultaneous} that
samples very small $32 \times 32$ patches from the original images. In
our experiments, we sample $224 \times 224$ pixel images from original
images varying from $300$ to $700$ pixels.

\section{Crowd Counting by Learning to Rank}
\label{sec:counting-loss}

In this section, we adapt the proposed framework to the task of crowd
counting. Perspective distortion, clutter, occlusion, non-uniform
distribution of people, complex illumination, scale variation, and a
host of other scene-incidental imaging conditions render person
counting and crowd density estimation in unconstrained images an
daunting problem.

Techniques for crowd counting have been recently improved using
Convolutional Neural Networks (CNNs). As with most CNN architectures,
however, these person counting and crowd density estimation techniques
are highly data-driven. For person counting, the labeling burden is
even more onerous than usual. Training data for person counting
requires that each individual person be meticulously labeled in
training images. It is for this reason that person counting and crowd
density estimation datasets tend to have only a few hundred images
available for training. As a consequence, crowd counting networks are
expected to benefit from additional unlabeled data which is used to
train a ranking proxy task.

\subsection{Crowd counting datasets}

\begin{figure*}[tpb]
\centering
\subfigure{\includegraphics[width=0.95\textwidth]{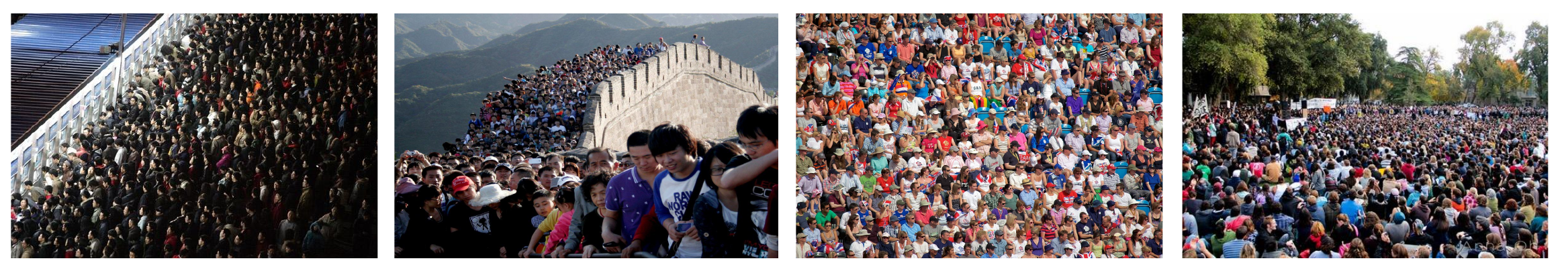}}
\subfigure{\includegraphics[width=0.95\textwidth]{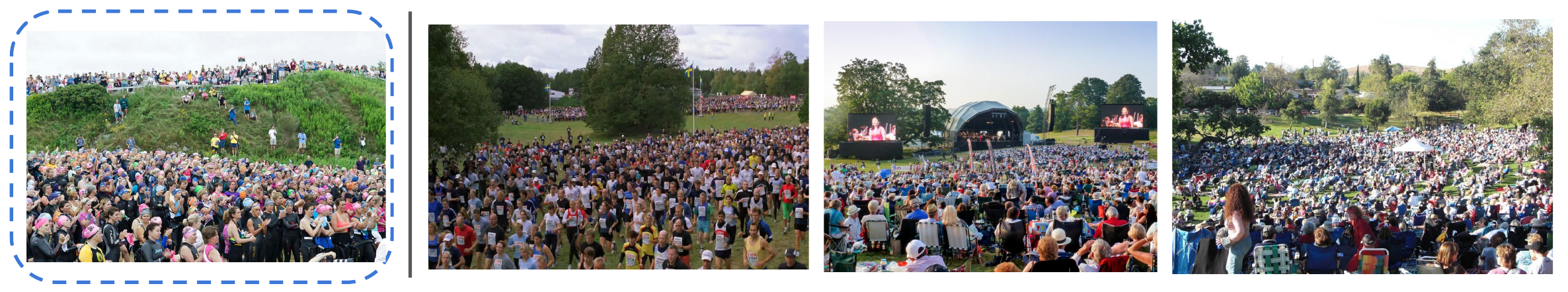}}
\caption{Example images from the retrieved crowd scene dataset. (top)
  Representative images using key words as query. (bottom)
  Representative images using training image as query image (the query
  image is depicted on the left).}
  \label{fig:data1}
\end{figure*}

We use two standard benchmark crowd counting datasets: 
\begin{itemize}[leftmargin=*]
\setlength\itemsep{0em}
\item \textbf{UCF\_CC\_50}~\cite{idrees2013multi}: This dataset
  contains 50 annotated images of different resolutions, illuminations
  and scenes. The variation of densities is very large among images
  from 94 to 4543 persons with an average of 1280 persons per image.

\item \textbf{ShanghaiTech}~\cite{zhang2016single}: Consists of 1198
  images with 330,165 annotated heads. This dataset includes two
  parts: 482 images in Part\_A which are randomly crawled from the
  Internet, and 716 images in Part\_B which are taken from busy
  streets. Both parts are further divided into training and evaluation
  sets. The training and test of Part\_A has 300 and 182 images,
  respectively, whereas that of Part\_B has 400 and 316 images,
  respectively.
\end{itemize}
Ground truth annotations for crowd counting typically consist of a set
of coordinates which indicate the 'center' (typically head center of a
person). To convert this data to crowd density maps we place a
Gaussian with standard deviation of 15 pixels and sum these for all
persons in the scene to obtain $y_i$. This is a standard procedure and
is also used in~\cite{onoro2016towards,zhang2016single}.

\subsection{Generating ranked image sets for counting}

\begin{figure}
  \begin{tabular}{c}
  \includegraphics[width=\columnwidth]{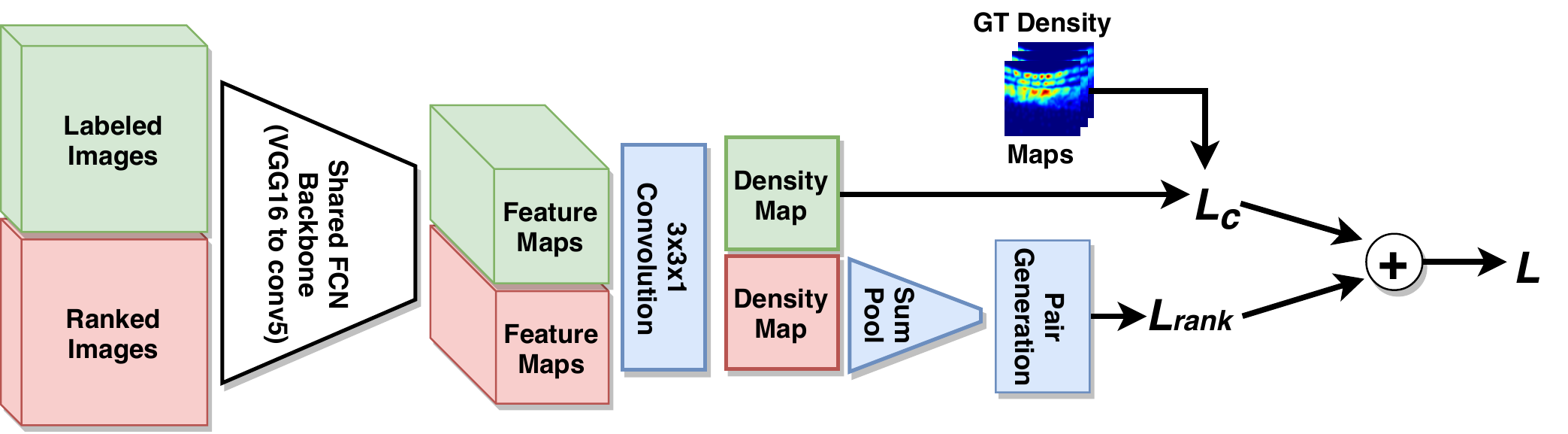} \\
  \includegraphics[width=\columnwidth]{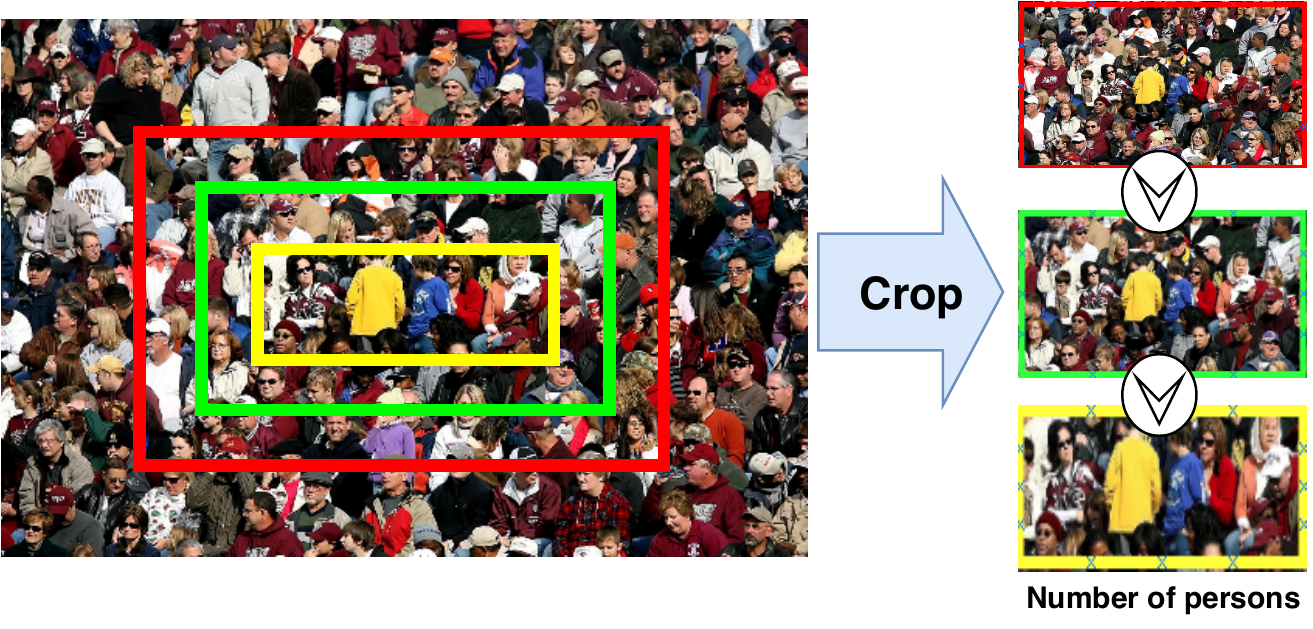}
 \end{tabular}
 \caption{
   Network architecture and ranked pair generation for crowd
   counting. \textbf{Top}: our counting network uses a VGG16 network truncated
   at the fifth convolutional layer (before maxpooling). To this
   network we add a $3 \times 3 \times 1$ convolutional layer with
   stride 1 which should estimate local crowd density. A sum pooling
   layer is added to the ranking channel of the network to arrive at a
   scalar value whose relative rank is known. \textbf{Bottom}: image
   pairs with known relative ranks are generated by selectively
   cropping unlabeled crowd images so that successive crops are
   entirely contained in previous ones.
 }
  \label{fig:counting-datagen}

\end{figure}

Here we show how to automatically generate the rankings from unlabeled
crowd counting images. The main idea is based on the observation that
all patches contained within a larger patch must have a fewer or equal
number of persons than the larger one (see
Fig.~\ref{fig:counting-datagen}). This observation allows us to
collect large datasets of crowd images with known relative ranks.
Rather than having to painstakingly annotate each person we are only
required to verify if the image contains a crowd. Given a crowd image
we extract ranked patches according to
Algorithm~\ref{table:algorithm2}.

To collect a large dataset of crowd images from the Internet we use
two different approaches:
\begin{itemize}
\item \textbf{Keyword query:} We collect a crowd scene dataset from
  Google Images by using different key words: \emph{Crowded,
    Demonstration, Train station, Mall, Studio, Beach}, all of which
  have high likelihood of containing a crowd scene. Then we delete
  images not relevant to our problem by simple visual inspection. In
  the end, we collected 1,180 high resolution crowd scene images,
  which is about 24x the size of the UCF\_CC\_50 dataset, 2.5x the
  size of ShanghaiTech Part\_A, and 2x the size of ShanghaiTech
  Part\_B. Note that \emph{no other annotation of images is
    performed.} Example images from this dataset are given in
  Fig.~\ref{fig:data1} (top row).

\item \textbf{Query-by-example image retrieval:} For each annotated
  benchmark dataset, we collect an unlabeled dataset using the
  training images as queries with the \emph{Google Images} visual
  image search engine. We choose the first ten similar images and
  remove irrelevant ones. For UCF\_CC\_50 we collected 256 images, for
  ShanghaiTech Part\_A 2229 images, and for ShanghaiTech Part\_B 3819
  images. An example of images returned for a specific query image is
  given in Fig.~\ref{fig:data1} (bottom row).
\end{itemize}

\begin{algorithm}[tb]
\centering
\begin{tabularx}{\columnwidth}{rp{0.75\columnwidth}}
\hline
\textbf{Input:}  & A crowd scene image, number of patches $k$ and scale factor $s$. \\
\textbf{Step 1:} & Choose an anchor point randomly from the anchor
region. The anchor region is defined to be $1/r$ the size of the original image, centered at the original image center, and with the same aspect ratio as the original image. \\
\textbf{Step 2:} & Find the largest square patch centered at the
anchor point and contained within the image boundaries. \\
\textbf{Step 3:} & Crop $k-1$ additional square patches, reducing size
iteratively by a scale factor $s$. Keep all patches centered at
anchor point. \\
\textbf{Step 4:} & Resize all $k$ patches to input size of network.\\
\textbf{Output:} & A list of patches ordered according to the number of persons in the patch.\\  \hline
\end{tabularx}
\caption{: Algorithm to generate ranked datasets.}
\label{table:algorithm2}
\end{algorithm}

\subsection{Crowd density estimation network}
We first explain the network architecture which is trained on
available crowd counting datasets with ground truth annotations (see Fig.~\ref{fig:counting-datagen}) . This
network regresses to a crowd density image which indicates the number
of persons per pixel (examples of such maps are given in
Fig.~\ref{fig:den}). A summation of all values in such a crowd density
image gives an estimate of the number of people in the scene. In the
experimental section we consider this network as the baseline method
to which we compare.

Our baseline network is derived from VGG-16~\cite{simonyan2014very}
pre-trained on ImageNet. VGG-16 consists of 13 convolutional layers
followed by three fully connected layers. We adapt the network to
regress to person density maps by removing its three fully connected
layers and the last max-pooling layer (pool5) to prevent further
reduction of spatial resolution. In their place we add a single
convolutional layer (a $3 \times 3 \times 512$ filter with stride 1
and zero padding to maintain same size) which directly regresses to
the crowd density map. As the counting loss $L_c$ we use the Euclidean
distance between the estimated and ground truth density maps, as given
by:
\begin{equation}
L_c = \frac { 1 }{N } \sum _{ i=1 }^{ N } (y_{i}-\hat{y}_{i})^2
\end{equation}
where $N$ is the number of images in a training batch, $y_i$ is ground
truth person density map of the $i$-th image in the batch, and the
prediction from the network as
$\hat{y}_i$. 

To further improve the performance of our baseline network, we
introduce multi-scale sampling from the available labeled datasets
during training. Instead of using the whole image as an input, we
randomly sample square patches of varying size (from 56 to 448
pixels). In the experimental section we verify that this multi-scale
sampling is important for good performance. Since we are processing
patches rather than images we use $\hat{y}_{i}$ to refer to the
estimate of patch $i$ from now on. The importance of multi-scale
processing of crowd data was also noted
in~\cite{boominathan2016crowdnet}.

Finally, we add a summation layer to the network. This summation layer
takes as an input the estimated density map and sums it to a single
number (the estimate of the number of persons in the image). This
output is used to compute the ranking loss (see
Eq.~(\ref{eq:ranking})) for the unlabeled images in the ranked
dataset. With respect to the basic architecture in
Fig.~\ref{fig:ranking_examples}, we use a sum polling layer as a
ranking specific layer as shown in Fig.~\ref{fig:counting-datagen}
(top).

\section{Experimental results}
\label{sec:exp}

In this section we report on an extensive set of experiments performed
to evaluate the effectiveness of learning from rankings for Image
Quality Assessment in section~\ref{sec:iqa-experiments} and crowd
counting in section~\ref{sec:counting-experiments}. We use the
Caffe~\cite{jia2014caffe} deep learning framework in all the
experiments.
\subsection{Image Quality Assessment (IQA)}
\label{sec:iqa-experiments}
We performed a range of experiments designed to evaluate the
performance of our approach with respect to baselines and the
state-of-the-art in IQA. These experiments make use of standard
datasets (for benchmark evaluation) and an additional dataset used for
generating ranked distorted image pairs as explained in
section~\ref{sec:iqa-loss}.

We use Stochastic Gradient Descent (SGD) with an initial learning rate
of 1e-4 for efficient Siamese network training and 1e-6 for
fine-tuning. Training rates are decreased by a factor of 0.1 every
10,000 iterations for a total of 50,000 iterations. For both training
phases we use $\ell_2$ weight decay (weight 5e-4). For multi-task
training, we use $\lambda=1$ in Eq.~(\ref{eq:multitask_loss}) with a
learning rate of 1e-6 on the LIVE dataset and 1e-5 on
TID2013. We found $\lambda=1$ to work well in initial
  experiments, but cross validating $\lambda$ on held-out data is
  expected to improve results for specific datasets. During training
we sample a single subimage from each training image per epoch. When
testing, we randomly sample 30 sub-images from the original images, as
suggested in~\cite{bianco2016use}, and pass all the trained
models. The average of all outputs of the sub-regions is the final
score for each distorted image.

Two evaluation metrics are traditionally used to evaluate the
performance of IQA algorithms: the Linear Correlation Coefficient
(LCC) and the Spearman Rank Order Correlation Coefficient (SROCC). LCC
is a measure of the linear correlation between the ground truth and
the predicted quality scores. Given $N$ distorted images, the ground
truth of $i$-th image is denoted by $y_i$, and the predicted score
from the network is $\hat{y}_i$. The LCC is computed as:
\begin{equation}
LCC = \frac{\sum_{ i=1 }^{ N }  ( { y }_{ i }-\overline { y } ) ( { \hat{y} }_{ i }-\overline { \hat{y} } )}
           {\sqrt{\sum _i^N ( { y }_{ i }-\overline { y } )^{ 2 } } \sqrt{ \sum _{ i }^{ N } ( \hat{ y }_{ i }-\overline { \hat{y} })^2  }}
\end{equation}
where $\overline { y }$ and $\overline { \hat{y} }$ are the means of
the ground truth and predicted quality scores, respectively.

Given $N$ distorted images, the SROCC is computed as:
\begin{equation}
  SROCC = 1 - \frac { 6\sum _{ i=1 }^{ N }{ { \left( { v }_{ i }-{ p }_{ i } \right)  }^{ 2 } }  }{ N\left( { N }^{ 2 }-1 \right)  },
\end{equation}
where $v_i$ is the \emph{rank} of the ground-truth IQA score $y_i$ in
the ground-truth scores, and $p_i$ is the \emph{rank} of $\hat{y}_i$
in the output scores for all $N$ images. The SROCC measures the
monotonic relationship between ground-truth and estimated IQA.

\begin{table}[tb]
\centering
\caption{Ablation study on the entire TID2013 database.}
\label{tab:iqa_ablation}
\begin{tabular}{l|cc}
\hline
Method              & LCC            & SROCC          \\ \hline
Baseline            & 0.663          & 0.612          \\ 
RankIQA             & 0.566          & 0.623 \\ \hline
 RankIQA+FT (Random)          & 0.775        & 0.738  \\

RankIQA+FT (Hard)        &   0.782    & 0.748  \\

RankIQA+FT (Ours)       &0.799     & 0.780        \\ \hline
MT-RankIQA (Random) & 0.802          & 0.770          \\
MT-RankIQA (Hard)   & 0.810          & 0.779          \\
MT-RankIQA (Ours)   & \textbf{0.827} & \textbf{0.806} \\ \hline
\end{tabular}
\end{table}

\begin{table*}[tb]
\centering
\caption{Performance evaluation (SROCC) on the entire TID2013 database.}
\label{tab:iqa_tid}

\resizebox{1.9\columnwidth}{!}{%
\begin{tabular}{c|ccccccccccccc}
\hline
Method                               & \#01                               & \#02                               & \#03                               & \#04                               & \#05                               & \#06                               & \#07                               & \#08                               & \#09                               & \#10                               & \#11                                & \#12                      & \#13                               \\ \hline
BLIINDS-II~\cite{saad2012blind}                           & 0.714                              & 0.728                              & 0.825                              & 0.358                              & 0.852                              & 0.664                              & 0.780                              & 0.852                              & 0.754                              & 0.808                              & 0.862                               & 0.251                     & 0.755                              \\
BRISQUE~\cite{mittal2012no}                              & 0.630                              & 0.424                              & 0.727                              & 0.321                              & 0.775                              & 0.669                              & 0.592                              & 0.845                              & 0.553                              & 0.742                              & 0.799                               & 0.301                     & 0.672                              \\
CORNIA-10K ~\cite{ye2012unsupervised}                           & 0.341                              & -0.196                             & 0.689                              & 0.184                              & 0.607                              & -0.014                             & 0.673                              & \textbf{0.896}                              & 0.787                              & 0.875                              & 0.911                               & 0.310                     & 0.625                              \\
HOSA   ~\cite{xu2016blind}                           & 0.853                              & 0.625                              & 0.782                              & 0.368                              & \textbf{0.905}                              & 0.775                              & 0.810                              & 0.892                     & 0.870                              & 0.893                              & \textbf{0.932}                               & \textbf{0.747}            & 0.701                              \\ \hline
RankIQA~\cite{liu2017rankiqa}    & \textbf{0.891} & \textbf{0.799} & \textbf{0.911} & \textbf{0.644} & 0.873 & \textbf{0.869} &\textbf{0.910} & 0.835          & \textbf{0.894} & \textbf{0.902} & 0.923  & 0.579 & 0.431          \\
RankIQA+FT~\cite{liu2017rankiqa}    & 0.667          & 0.620          & 0.821          & 0.365          & 0.760          & 0.736          & 0.783          & 0.809          & 0.767          & 0.866          & 0.878           & 0.704 & 0.810 \\  
MT-RankIQA     & 0.780     & 0.658         & 0.882        & 0.424       & 0.839      & 0.762        & 0.852       & 0.861       & 0.799        & 0.879        & 0.909        & 0.744  & \textbf{0.824} \\ \hline
Method                               & \#14                               & \#15                               & \#16                               & \#17                               & \#18                               & \#19                               & \#20                               & \#21                               & \#22                               & \#23                               & \multicolumn{1}{c|}{\#24}          & \multicolumn{2}{c}{ALL}                                        \\ \hline

BLIINDS-II     ~\cite{saad2012blind}                       & 0.081                              & 0.371                              & 0.159                              & -0.082                             & 0.109                              & 0.699                              & 0.222                              & 0.451                              & 0.815                              & 0.568                              & \multicolumn{1}{c|}{0.856}          & \multicolumn{2}{c}{0.550}                                      \\
BRISQUE   ~\cite{mittal2012no}                            & 0.175                              & 0.184                              & 0.155                              & 0.125                              & 0.032                              & 0.560                              & 0.282                              & 0.680                              & 0.804                              & 0.715                              & \multicolumn{1}{c|}{0.800}          & \multicolumn{2}{c}{0.562}                                      \\
CORNIA-10K    ~\cite{ye2012unsupervised}                       & 0.161                              & 0.096                              & 0.008                              & 0.423                              & -0.055                             & 0.259                              & 0.606                              & 0.555                              & 0.592                              & 0.759                              & \multicolumn{1}{c|}{0.903}          & \multicolumn{2}{c}{0.651}                                      \\
HOSA  ~\cite{xu2016blind}                                & 0.199                              & 0.327                              & 0.233                              & 0.294                              & 0.119                              & 0.782                              & 0.532                              & 0.835                              & \textbf{0.855}                     & \textbf{0.801}                     & \multicolumn{1}{c|}{\textbf{0.905}} & \multicolumn{2}{c}{0.728}                                      \\ \hline
RankIQA~\cite{liu2017rankiqa}   & 0.463          & \textbf{0.693} & \textbf{0.321} & \textbf{0.657} & 0.622          & \textbf{0.845} & 0.609          & \textbf{0.891} & 0.788          & 0.727          & \multicolumn{1}{c|}{0.768}          & \multicolumn{2}{c}{0.623}                                      \\
RankIQA+FT~\cite{liu2017rankiqa}                          & \textbf{0.512}                     & 0.622                              & 0.268                              & 0.613                              & 0.662                     & 0.619                              & 0.644                     & 0.800                              & 0.779                              & 0.629                              & \multicolumn{1}{c|}{0.859}          & \multicolumn{2}{c}{0.780}                             \\ 


MT-RankIQA            & 0.458                     & 0.658                              & 0.198                             & 0.554                              & \textbf{0.669}                     & 0.689                              & \textbf{0.760}                     & 0.882                             & 0.742                             & 0.645                              & \multicolumn{1}{c|}{0.900}          & \multicolumn{2}{c}{\textbf{0.806}}                             \\ \hline
\end{tabular} }

\end{table*}

\subsubsection{Ablation study}

In this experiment, we evaluate the effectiveness of using rankings to
estimate image quality. We compare our multi-task approach with
different baselines: fine-tuning the VGG-16 network initialized from
ImageNet to obtain the mapping from images to their predicted scores
(which we call Baseline in our experiments), and two other baselines
from our previous work~\cite{liu2017rankiqa}: VGG-16 (initialized
pre-trained ImageNet weights) trained on ranking data (called
RankIQA), and our RankIQA approach fine-tuned on TID2013 after
training using ranked pairs of images (called RankIQA+FT). In order to
evaluate the effectiveness of our sampling method in the multi-task
setting, we also report the accuracy for multi-task training (called
MT-RankIQA) with different sampling methods: standard random pair
sampling, and a hard-negative mining method similar
to~\cite{simo2015discriminative}. 
For standard random pair sampling we randomly choose 36 pairs for each
mini-batch from the training sets. For the hard negative mining
strategy we start from 36 pairs in a mini-batch, and gradually
increase the number of hard pairs every 5000 iterations. For our
method we pass 72 images in each mini-batch. With these settings the
computational costs for all three methods are equal, since at each
iteration 72 images are passed through the network.

We follow the experimental protocol used in
HOSA~\cite{xu2016blind}. The entire TID2013 database including all
types of distortions is divided into 80\% training images and 20\%
testing images according to the reference images and their distorted
versions. The results are shown in Table~\ref{tab:iqa_ablation}, where
ALL means testing all distortions together. All the experiments are
performed 10 times and the average SROCC is
reported

From Table~\ref{tab:iqa_ablation}, we can draw several
conclusions. First, it is hard to obtain good results by training a
deep network directly on IQA data. This is seen in the Baseline
results and is due to the scarcity of training
data. 
Second, competitive results are obtained using RankIQA without access
to the ground truth of IQA dataset during training the ranking
network, which strongly demonstrates the effectiveness of training on
ranking data. The RankIQA-trained network alone does not provide
accurate IQA scores (since it has never seen any) but does yield high
correlation with the IQA scores as measured by SROCC. After
fine-tuning on the TID2013 database, we considerably improve
performance for all sampling methods: with random sampling we improve
on the baseline by 11\%, while our efficient sampling method further
outperforms random sampling by 2.4\% in terms of LCC (similar
conclusions can be drawn from the SROCC results.

Finally, in Table~\ref{tab:iqa_ablation} we also compare the three
optimization methods for multi-task training: random pair sampling,
hard negative mining, and our proposed efficient Siamese
backpropagation. We see that our efficient back-propagation strategy
with multi-task setting obtains the best results, further improving
the accuracy by 2.8\% on LCC and 2.6\% on SROCC with respect to
RankIQA+FT. This clearly shows the benefits of the proposed
backpropagation scheme. In the next section, we use MT-RankIQA to
refer to our method trained with the multi-task loss using our
efficient Siamese backpropagation method.

To demonstrate the ability of our approach to generalize to
  unseen distortions, we trained our multi-task approach with
  different numbers of synthetic distortions as auxiliary data
  combined with all labeled data in the TID2013 dataset. All results are the average of three runs. As shown in
  Table~\ref{tab:unseen}, adding more distortions results in increased overall
  performance on the test set. Note also that adding specific
  distortions not only consistently improves accuracy on seen
  distortions consistently, but also on unseen ones.
 
\begin{table}[tb]
\centering
\caption{Generalization to unseen distortions. Results of MT-RankIQA with different numbers of synthetic distortions as auxiliary data combined with all labeled data in the TID2013 dataset. Results show that also on the unseen distortions a significant gain in performance is obtained. }
\label{tab:unseen}

\begin{tabular}{cccc}
\hline
\multicolumn{2}{c}{\multirow{2}{*}{Overall LCC accuracy}} & \multicolumn{2}{c}{Average gain}      \\
\multicolumn{2}{c}{}                                      & Seen distortions & Unseen distortions \\ \hline
Baseline                         & 0.687                  & --                & --                  \\
7 distortions                    & 0.803                  & +0.102           & +0.016             \\
11 distortions                   & 0.817                  & +0.102           & +0.032             \\
15 distortions                   & 0.823                  & +0.104           & +0.067             \\
All                              & 0.829                  & +0.109           & +0.075             \\ \hline
\end{tabular}
\end{table}

\subsubsection{Comparison with the state-of-the-art}

We compare the performance of our method with state-of-the-art
Full-reference IQA (FR-IQA) and NR-IQA methods on both TID2013 and
LIVE dataset.

\begin{table}
\begin{center}
\caption{LCC (above) and SROCC (below) evaluation on  LIVE
  dataset. We divide approaches into full-reference (FR-) and
no-reference (NR-) IQA.}
\label{tab:iqa_live}
\resizebox{0.95\columnwidth}{!}{%
\begin{tabular}{|c|c|c|c|c|c|c|c|}
\hline
 & \textbf{LCC} & \textbf{JP2K}  & \textbf{JPEG}  & \textbf{GN} & \textbf{GB} & \textbf{FF} & \textbf{ALL} \\ \hline
\parbox[t]{2mm}{\multirow{4}{*}{\rotatebox[origin=c]{90}{\textbf{FR-IQA}}}}
& PSNR                             & 0.873 & 0.876 & 0.926 & 0.779 & 0.87  & 0.856 \\
& SSIM~\cite{wang2004image}        & 0.921 & 0.955 & 0.982 & 0.893 & 0.939 & 0.906 \\
& FSIM~\cite{zhang2011fsim}        & 0.91  & 0.985 & 0.976 & 0.978 & 0.912 & 0.96  \\
& DCNN~\cite{liang2016image}       & --  & -- & -- & -- & -- & 0.977  \\ \hline
\parbox[t]{2mm}{\multirow{8}{*}{\rotatebox[origin=c]{90}{\textbf{NR-IQA}}}}
& DIVINE~\cite{moorthy2011blind}   & 0.922 & 0.921 & 0.988 & 0.923 & 0.888 & 0.917 \\
& BLIINDS-II~\cite{saad2012blind}  & 0.935 & 0.968 & 0.98  & 0.938 & 0.896 & 0.93  \\
& BRISQUE~\cite{mittal2012no}      & 0.923 & 0.973 & 0.985 & 0.951 & 0.903 & 0.942 \\
& CORNIA~\cite{ye2012unsupervised} & 0.951 & 0.965 & 0.987 & 0.968 & 0.917 & 0.935 \\
& CNN~\cite{kang2014convolutional} & 0.953 & 0.981 & 0.984 & 0.953 & 0.933 & 0.953 \\
& SOM~\cite{zhang2015som}          & 0.952 & 0.961 & 0.991 & 0.974 & 0.954 & 0.962  \\
& DNN~\cite{bosse2016deep}          & -- & -- & -- & -- & -- & 0.972 \\ \cline{2-8}

& \textbf{MT-RankIQA}  & \textbf{0.972} & \textbf{0.978} & \textbf{0.988} & \textbf{0.982} & \textbf{0.971} & \textbf{0.976} \\
\hline
\hline
\hline
& \textbf{SROCC} & \textbf{JP2K}  & \textbf{JPEG}  & \textbf{GN} & \textbf{BLUR}  & \textbf{FF}    & \textbf{ALL}   \\ \hline
\parbox[t]{2mm}{\multirow{4}{*}{\rotatebox[origin=c]{90}{\textbf{FR-IQA}}}}
& PSNR                             & 0.87  & 0.885 & 0.942 & 0.763 & 0.874 & 0.866 \\
& SSIM~\cite{wang2004image}        & 0.939 & 0.946 & 0.964 & 0.907 & 0.941 & 0.913 \\
& FSIM~\cite{liang2016image}       & 0.97  & 0.981 & 0.967 & 0.972 & 0.949 & 0.964 \\
& DCNN~\cite{liang2016image}       & --  & -- & -- & -- & -- & 0.975 \\ \hline
\parbox[t]{2mm}{\multirow{8}{*}{\rotatebox[origin=c]{90}{\textbf{NR-IQA}}}}
& DIVINE~\cite{moorthy2011blind}   & 0.913 & 0.91  & 0.984 & 0.921 & 0.863 & 0.916 \\
& BLIINDS-II~\cite{saad2012blind}  & 0.929 & 0.942 & 0.969 & 0.923 & 0.889 & 0.931 \\
& BRISQUE~\cite{mittal2012no}      & 0.914 & 0.965 & 0.979 & 0.951 & 0.887 & 0.94  \\
& CORNIA~\cite{ye2012unsupervised} & 0.943 & 0.955 & 0.976 & 0.969 & 0.906 & 0.942 \\
& CNN~\cite{kang2014convolutional} & 0.952 & 0.977 & 0.978 & 0.962 & 0.908 & 0.956 \\
& SOM~\cite{zhang2015som}          & 0.947 & 0.952 & 0.984 & 0.976 & 0.937 & 0.964 \\
&DNN~\cite{bosse2016deep}          & -- & -- & -- & -- & -- & 0.960 \\ \cline{2-8}

& \textbf{MT-RankIQA}  & \textbf{0.971} & \textbf{0.978} & \textbf{0.985} & \textbf{0.979} &  \textbf{0.969} & \textbf{0.973} \\ \hline
\end{tabular}}
\end{center}
\end{table}

\minisection{Evaluation on TID2013.} Table~\ref{tab:iqa_tid} includes
results of state-of-the-art methods. We see that for several very
challenging distortions (14 to 18), where all other methods fail, we
obtain satisfactory results. For individual distortions, there is a
huge gap between RankIQA and other methods on most distortions. The
state-of-the-art method HOSA performs slightly better than our methods
on 6 out of 24 distortions. For all distortions, RankIQA+FT achieves
about 5\% higher than HOSA, and about 3\% more is gained by using
multi-task training. Our methods also perform well on distortions for
which were unable to generate rankings. This indicates that different
distortions share some common representation and training the network
jointly on all distortions also improves results for the distortions
for which we did not generate rankings.  

\minisection{Evaluation on LIVE.} As done
in~\cite{kang2014convolutional,zhang2015som}, we randomly split the
reference images and their distorted version from LIVE into 80\%
training and 20\% testing sample and compute the average LCC and SROCC
scores on the testing set after training to convergence. This process
is repeated ten times and the results are averaged. These results are
shown in Table~\ref{tab:iqa_live}. For fair comparison with the
state-of-the-art, we train our ranking model on four distortions (all
but FF), but we fine-tune our model on all five distortions in the
LIVE dataset to compute ALL. As shown in Table~\ref{tab:iqa_live} our
approach improves by 0.4\% and 1\% in LCC and SROCC, respectively, the
best NR-IQA results reported on ALL distortions. This indicates that
our method outperforms existing work including the current
state-of-the-art NR-IQA method SOM~\cite{zhang2015som} and
DNN~\cite{bosse2016deep}, and even achieves competitive results as
state-of-the-art FR-IQA method DCNN~\cite{liang2016image} which, being
a full-reference approach, has the benefit of having access to the
high-quality original reference image.

\subsubsection{Active learning for IQA}
  
We demonstrate the effectiveness of active learning on the IQA
  problem (see Algorithm~\ref{table:algorithm}). As a dataset we
  consider all 24 distortions for each of 19 reference images from  TID2013, yielding a dataset of 456 image-distortion pairs. Each image is distorted for each distortion at five distortion levels. During active learning we aim to select which images with what particular distortion are expected to most improve performance. We add all five distortion levels of the selected image with the particular distortion to the labeled pool. 
The active learning loop starts with 10\% of this data labeled; hence the
  labeled samples $D$ is 10\% and the examples $E$ consist of the
  remaining 90\% of the data without labels. We then perform $T=9$
  active learning cycles.  In each cycle $S=10\%$ images with a particular distortion are added
  incrementally, using the full training set from the previous
  iteration to estimate informativeness for the remaining unlabeled
  data. We use $K=100$ in the experiment, and compare results to a
  baseline of randomly adding 10\% additional labeled samples at each
  step.

\begin{figure}[tb]
\centering
\includegraphics[width = 7.0cm]{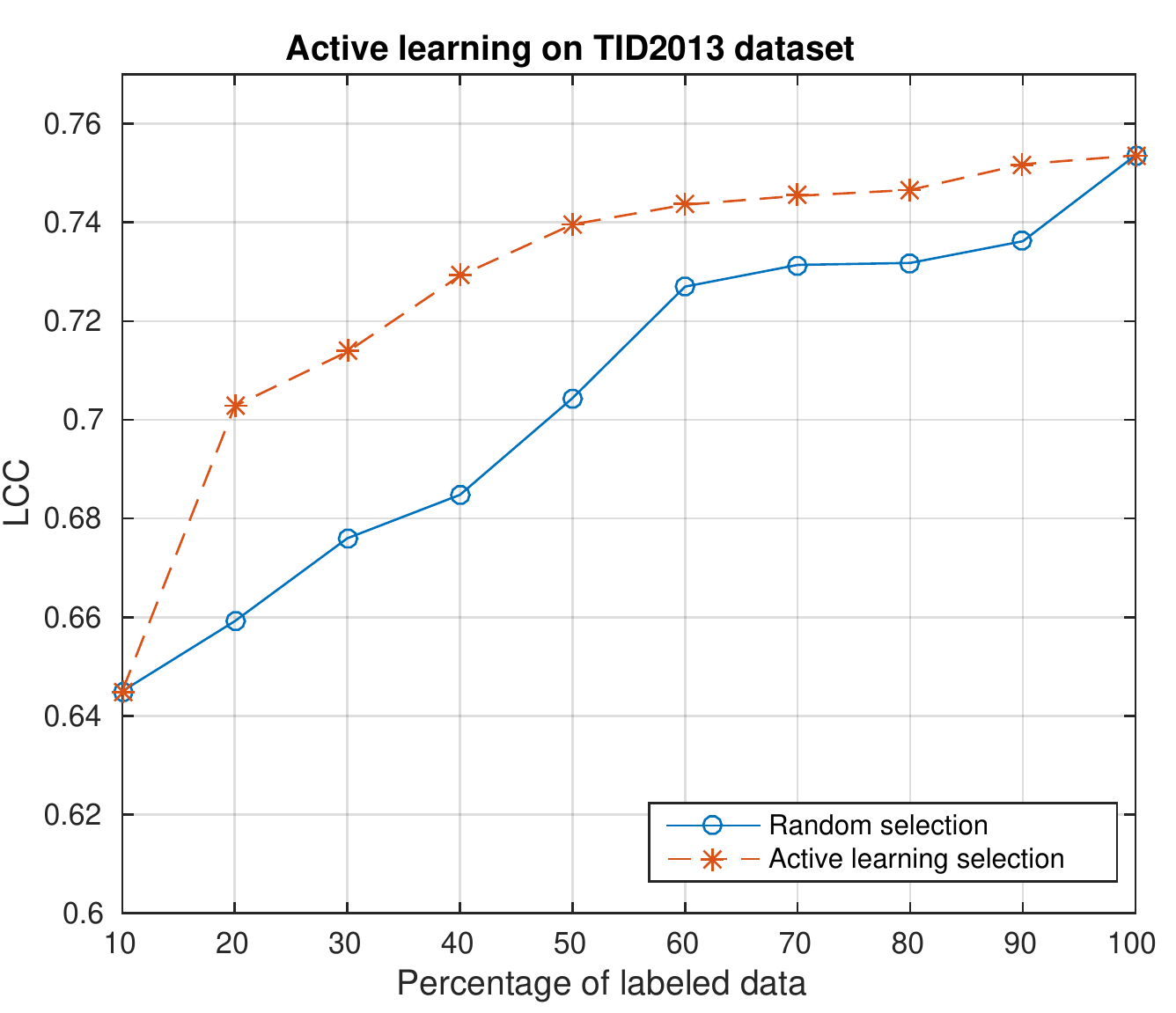}

\caption{Active learning results on TID 2013. We plot LCC as a
    function of the percentage of labeled data used from the training
    set.} 
  \label{fig:active_iqa2}
\end{figure}

Results for active learning are given in
  Fig.~\ref{fig:active_iqa2}. It is clear that our active learning
  algorithm obtains results superior to random selection. When adding
  an additional 10\% of data (using a total of 20\% labeled data) we
  achieve similar accuracy as adding an additional 40\% data (using a
  total of 50\% labeled data) with random selection, thereby reducing
  the labeling cost by 75\% compared to the baseline.

\subsection{Crowd counting}
\label{sec:counting-experiments}

Here we report on a range of experiments evaluating our
approach with respect to baselines and the state-of-the-art methods
for crowd counting.

We use SGD with a batch size of 25 for both ranking and counting, and
thus a batch size of 50 for multi-task training. For the ranking plus
fine-tuning method, the learning rate is 1e-6 for both ranking and
fine-tuning. For multi-task training, we found that $\lambda=100$
yielded good results on all datasets. Cross validating $\lambda$ on
held-out data is expected to improve results for specific
datasets. Learning rates are decreased by a factor of 0.1 every 5,000
iterations for a total of 10K iterations. For both training phases we
use $\ell_2$ weight decay with a weight of 5e-4. During training we
sample one sub-image from each training image per epoch. We perform
down-sampling of three scales and up-sampling of one scale on the
UCF\_CC\_50 dataset and only up-sampling of one scale on the
ShanghaiTech dataset. The number of ranked crops $k=5$, the scale
factor $s=0.75$, and the anchor region $r=8$ (see
Algorithm~\ref{table:algorithm2}).

Following existing work, we use the mean absolute error (MAE) and the
mean squared error (MSE) to evaluate different methods. These are
defined as follows:
\begin{eqnarray}
  MAE & = & \frac { 1 }{ N } \sum _{ i=1 }^{ N }| y_i - \hat{y}_i | \\
  MSE & = & \sqrt{\frac { 1 }{ N } \sum _{ i=1 }^{ N } ( y_i- \hat{y}_i)^2}
\end{eqnarray}
where $N$ is the number of test images, $y_i$ is the ground truth
number of persons in the $i$th image, and $\hat{y}_i$ is number of
persons predicted by the network in the $i$th image.

\begin{table}[tb]
\centering
\caption{Ablation study on UCF\_CC\_50 with five-fold cross
  validation. 
}
 \resizebox{1.0\columnwidth}{!}{%

\begin{tabular}{lccccc|c}
\hline
\textbf{Method}          & \textbf{Split 1} & \textbf{Split 2} & \textbf{Split 3} & \textbf{Split 4} & \textbf{Split 5} & \textbf{Ave MAE}        \\ \hline \hline
Basic CNN       & 701.41 & 394.52 & 497.57 & 263.56 & 415.23 & 454.45  \\
\: + Pre-trained model       & 570.01 & 350.63 & 334.89 & 184.79 & 202.41 & 328.54  \\
\: + multi-scale       & 532.85 & 307.43 & 266.75 & 216.96 & 216.35 & 308.06 \\  \hline
 Ranking+FT       & 552.68 & 375.38 &  241.28 & 211.66 & 247.70 & 325.73  \\ 
Multi-task (Random) & 462.71 & 345.31 & 218.71 & 226.44 & 210.19 & 292.67 \\ 
 Multi-task (Hard) & 460.35 & 343.91 & 208.23 & 221.75 & 205.57 & 287.96  \\ 

 Multi-task (Ours) & 443.68 & 340.31 & 196.76 & 218.48 & 199.54 & \textbf{279.60 }  \\ \hline

\end{tabular}}
\label{table:ablation}
\end{table}

\subsubsection{Ablation study.}
We begin with an ablation study on the UCF\_CC\_50 dataset. The aim is
to evaluate the relative gain of the proposed improvements and to
evaluate the use of a ranking loss against the baseline. The ranked
images in this experiment are generated from the Keyword dataset. The
results are summarized in Table~\ref{table:ablation}. We can
immediately observe the benefit of using a pre-trained ImageNet model
in crowd counting, with a significant drop in MAE of around 28\%
compared to the model trained from scratch. By using both multi-scale
data augmentation and starting from a pre-trained model, another
improvement of around 6\% is obtained.

The Ranking+FT method performs worse than directly fine-tuning from a
pre-trained ImageNet model. This is probably caused by the
poorly-defined nature of the self-supervised task. To optimize this
task the network could decide to count anything, e.g. `hats', `trees',
or `people with red shirts', or even just `edges' -- all of which
would satisfy the ranking constraints that are imposed.

Next, we compare the three sampling strategies for combining the
ranking and counting losses for multi-task training. When using
multi-task training with random pair sampling, the average MAE is
reduced by about 15 points. Hard mining obtains about 5 points average
MAE less than random sampling. However, our efficient back-propagation
approach reduces the MAE further to 279.6. This shows that by
\emph{jointly} learning both the self-supervised and crowd counting
tasks, the self-supervised task is forced to focus on counting
persons. Given its superior results, we consider only the
``Multi-task'' with efficient back-propagation approach for the
remainder of the experiments.
\begin{table}
\centering
\caption{MAE and MSE error on the UCF\_CC\_50 dataset.}
\begin{tabular}{rcc}
\hline
\textbf{Method} & \textbf{MAE}   & \textbf{MSE}   \\ \hline \hline
Idrees et al.  \cite{idrees2013multi}     & 419.5 & 541.6 \\
Cross-scene~ \cite{zhang2015cross}       & 467.0 & 498.5 \\
MCNN \cite{zhang2016single}      & 377.6 & 509.1 \\
Onoro et al.  \cite{onoro2016towards}      & 333.7 & 425.2 \\
Walach et al. \cite{walach2016learning}     & 364.4 & 341.4 \\
Switching-CNN~ \cite{Sam_2017_CVPR}     & 318.1 & 439.2 \\
CP-CNN~\cite{sindagi2017generating}       & 295.8 & \textbf{320.9} \\ 
ACSCP~ \cite{shen2018crowd}       & 291.0 & 404.6 \\
 CSRNet~\cite{li2018csrnet} & 266.1 & 397.5 \\ 
 ic-CNN~\cite{ranjan2018iterative} & \textbf{260.9} & 365.5 \\  \hline
 Multi-task (Query-by-example)   & 291.5 & 397.6 \\
 Multi-task (Keyword)   & 279.6 &  408.1 \\ \hline
\end{tabular}

\label{table:ucf}
\end{table}

\subsubsection{Comparison with the state-of-the-art}

\begin{figure*}[tpb]
\centering

\includegraphics[width=0.33\textwidth]{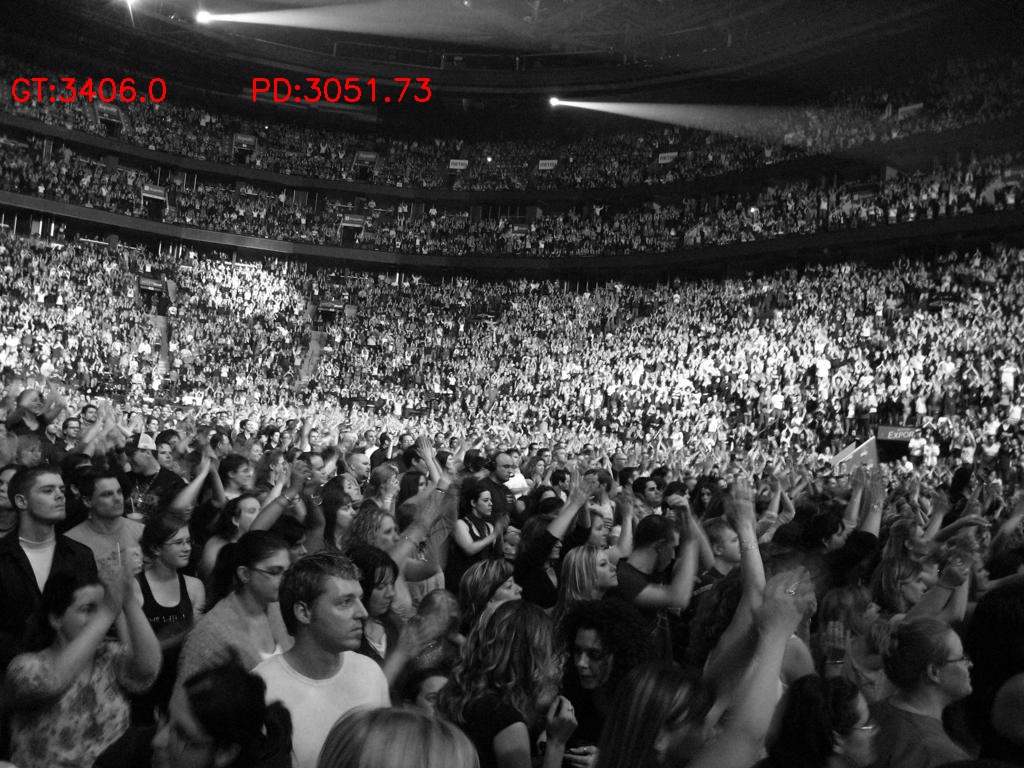} 
 \includegraphics[width=0.33\textwidth]{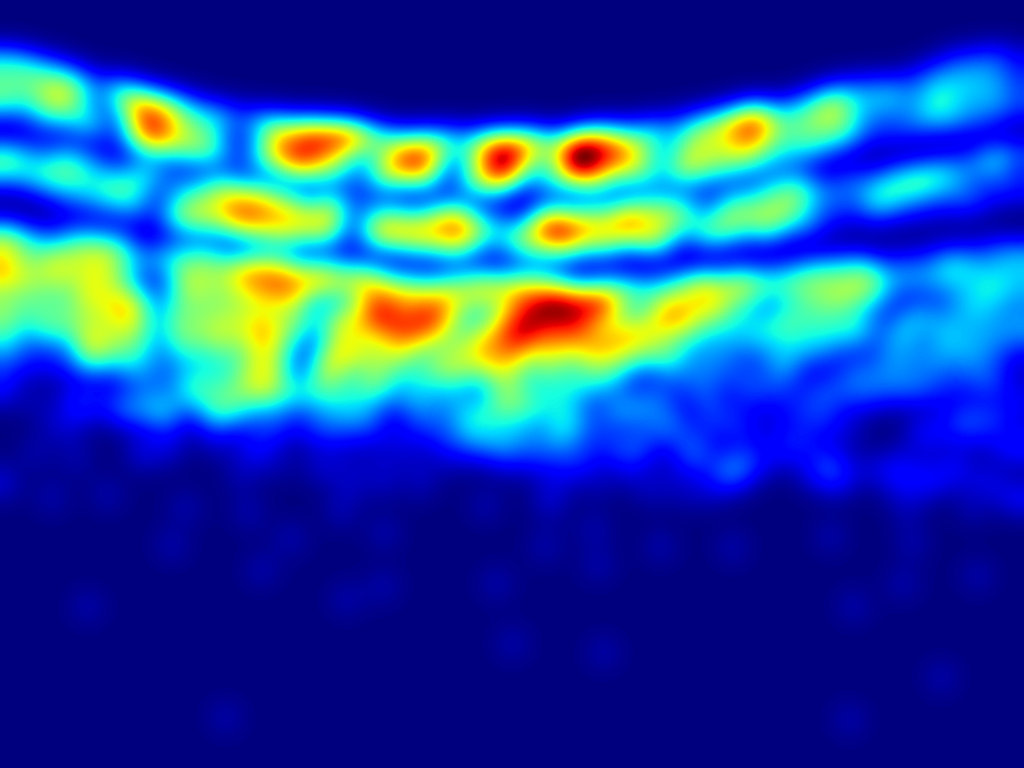}
 \includegraphics[width=0.33\textwidth]{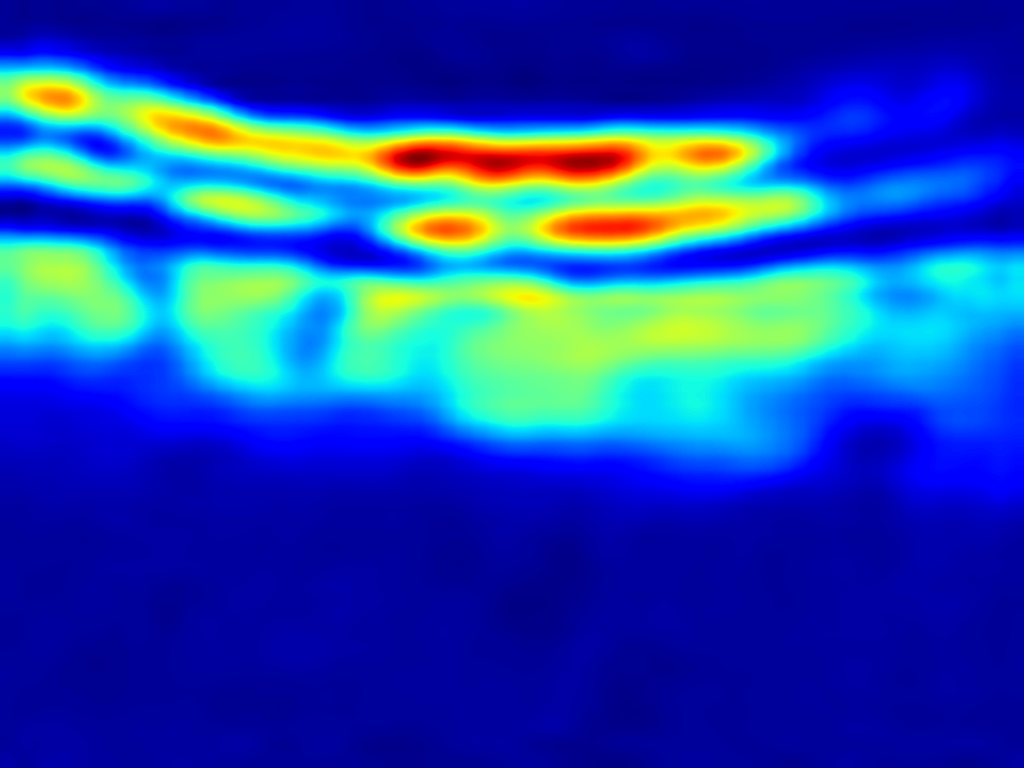}\\
 \includegraphics[width=0.33\textwidth]{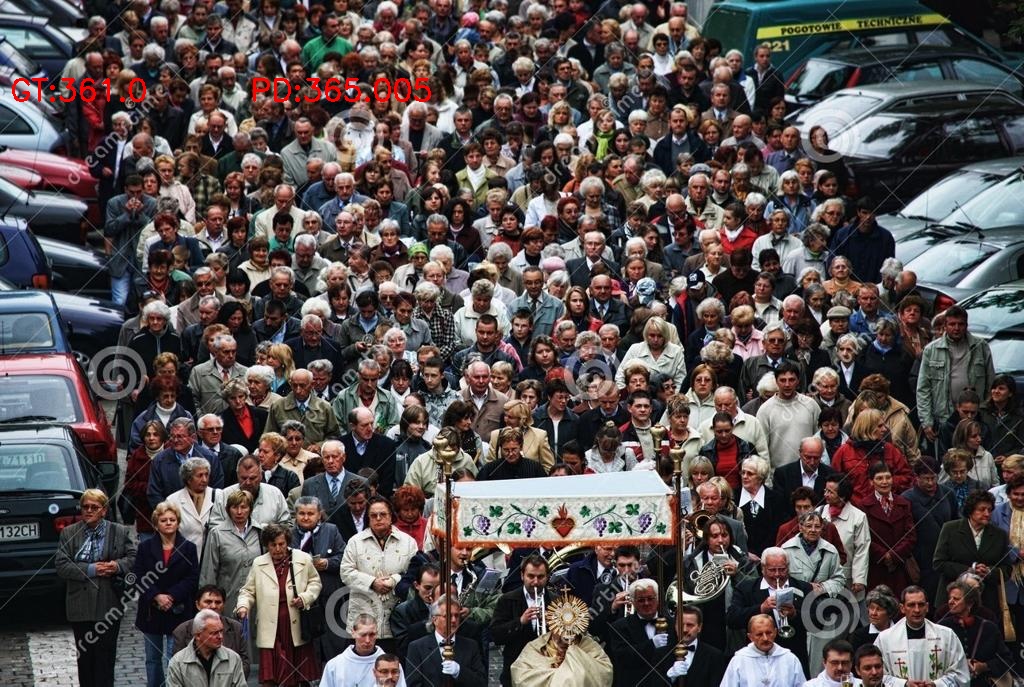}
 \includegraphics[width=0.33\textwidth]{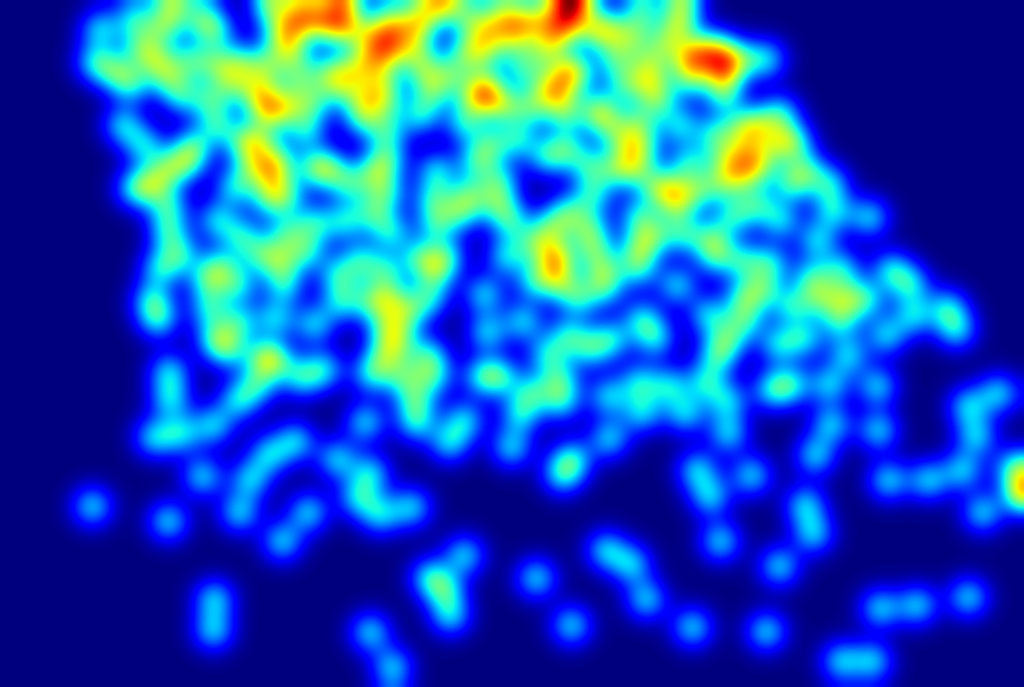}
 \includegraphics[width=0.33\textwidth]{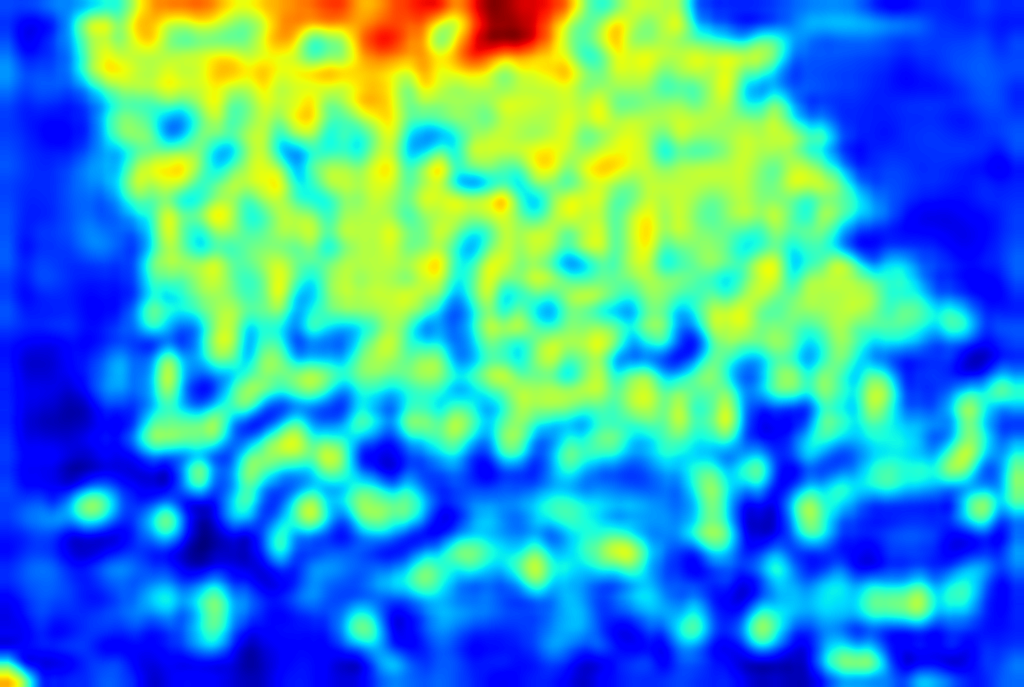}

\caption{Examples of predicted density maps for the UCF\_CC\_50 (Top row, true count: 3406 prediction: 3052) and ShanghaiTech datasets (Bottom row, true count: 361 prediction: 365).  Left column: crowd image. Middle column: ground truth. Right column: prediction.}
  \label{fig:den}
\end{figure*}

\minisection{Evaluation on the UCF\_CC\_50 dataset.}  A five-fold
cross-validation was performed for evaluating the methods. Results are
shown in Table~\ref{table:ucf}. 
Our multi-task training method using the unlabeled Keyword
  Dataset reduces the MAE from 291.0 to 279.6, which is comparable to
  ACSCP~\cite{shen2018crowd} which was published at the same time as
  our original work. Our approach performs slightly worse than the
  state-of-the-art methods CSRNet~\cite{li2018csrnet} and
  ic-CNN~\cite{ranjan2018iterative} (also published around the same
  time as our original work), but in general our model has fewer
  parameters than CSRNet and a simpler inference procedure compared to
  ic-CNN. However, the MSE of our method on UCF\_CC\_50 dataset is
worse than the state-of-the-art
methods~\cite{walach2016learning,sindagi2017generating,ranjan2018iterative},
but achieves competitive results compared
to~\cite{shen2018crowd,li2018csrnet}. This indicates that our method
and also \cite{shen2018crowd} work better in general but have more
extreme outliers. Compared to training on the Keyword dataset,
learning from the Query-by-example dataset is slightly worse, which
might be because most images from UCF\_CC\_50 are black and white with
low resolution, which often does not lead to satisfactory query
results. An example of prediction in UCF\_CC\_50 using our network is
shown in Fig.~\ref{fig:den}.

\begin{table}[tb]
\centering
\caption{MAE and MSE error on ShanghaiTech.}

\resizebox{\columnwidth}{!}{%
\begin{tabular}{r|rr|rr}
\hline
       & \multicolumn{2}{c|}{\textbf{Part A}} & \multicolumn{2}{c}{\textbf{Part B}} \\
\textbf{Method} & \textbf{MAE}          & \textbf{MSE}         & \textbf{MAE}          & \textbf{MSE}         \\ \hline \hline
Cross-scene~ \cite{zhang2015cross}     & 181.8        & 277.7       & 32.0         & 49.8        \\
MCNN~\cite{zhang2016single}     & 110.2        & 173.2       & 26.4         & 41.3        \\
Switching-CNN~\cite{Sam_2017_CVPR}    & 90.4         & 135.0       & 21.6         & 33.4        \\
CP-CNN~ \cite{sindagi2017generating}      & 73.6         & 106.4       & 20.1         & 30.1        \\ 
ACSCP~  \cite{shen2018crowd}        & 75.7         & \textbf{102.7}       & 17.2        & 27.4       \\
CSRNet~  \cite{li2018csrnet}  & \textbf{68.2}         & 115.0       & \textbf{10.6}        & \textbf{16.0}  \\ ic-CNN~\cite{ranjan2018iterative} & 68.5         & 116.2       & 10.7        & \textbf{16.0} \\ \hline
\textbf{Ours}: Multi-task (Query-by-example)   & 72.0        & 106.6 & 14.4         & 23.8      \\
\textbf{Ours}: Multi-task (Keyword)   & 73.6         & 112.0      & 13.7         & 21.4        \\ \hline
\end{tabular}
}
\label{table:shanghai}
\end{table}
\minisection{Evaluation on the ShanghaiTech dataset.} Looking at
Table~\ref{table:shanghai}, we can draw conclusions similar to those
on UCF\_CC\_50. We see here that using the Query-by-example
  Dataset further improves by about 2\% on ShanghaiTech -- especially
  for Part\_A, where our approach surpasses the state-of-the-art
  method~\cite{sindagi2017generating}, but is still slightly worse
  than CSRNet~\cite{li2018csrnet} and
  ic-CNN~\cite{ranjan2018iterative}. An example of prediction by our
network on ShanghaiTech is given in Fig.~\ref{fig:den}. For
comparison, we also provide the results of our baseline method
(including fine-tuning from a pre-trained model and multi-scale data
augmentation) on this dataset: $MAE=77.7$ and $MSE=115.9$ on Part A,
and $MAE=14.7$ and $MSE=24.7$ on Part
B.  

\minisection{Evaluation on the WorldExpo'10 dataset} The
  WorldExpo'10 dataset~\cite{zhang2015cross} consists of 3980 frames
  of size 576$\times$ 720 from 1132 video sequences captured by 108
  surveillance cameras. The dataset is split into training set with
  103 scenes and test set consisting of 5 different scenes. Regions of
  interest (ROIs) are provided for the whole dataset, which are used
  as a mask during testing. We consider training directly on the only
  ground truth labels as a baseline, and for our multi-task approach,
  when we test on one specific scene, the rest of scenes in the test
  set are used to generate the ranked image set. We compare our
  multi-task training to the baseline and other state-of-the-art
  methods in Table~\ref{table:worldexpo}. It is clear that our
  multi-task training approach outperforms the baseline in all five
  cases and achieves comparable results compared to other methods in
  terms of MAE.

\begin{table}[tb]
\centering
\caption{MAE results on the WorldExpo'10 dataset.}
 \resizebox{1.0\columnwidth}{!}{%
\begin{tabular}{rccccc|c}
  \hline
\textbf{Method}          & \textbf{Scene 1} & \textbf{Scene 2} & \textbf{Scene 3} & \textbf{Scene 4} & \textbf{Scene 5} & \textbf{Ave MAE}        \\ \hline \hline
MCNN ~\cite{zhang2016single}       & 3.4 & 20.6 & 12.9 & 13.0 & 8.1 & 11.6  \\
 Switching-CNN~\cite{Sam_2017_CVPR}       & 4.4 & 15.7 & 10.0 & 11.0 & 5.9 & 9.4  \\
 CP-CNN~ \cite{sindagi2017generating}       & 2.9 & 14.7 & 10.5 & 10.4 & 5.8 & 8.9 \\
 ACSCP~  \cite{shen2018crowd}     & 2.8 & 14.05 & 9.6 & 8.1 & 2.9 & \textbf{7.5}  \\
 
 CSRNet~  \cite{li2018csrnet} & 2.9   & 11.5 & 8.6 & 16.6 & 3.4 & 8.86\\ 
 ic-CNN~\cite{ranjan2018iterative}     & 17.0   & 12.3 & 9.2 & 8.1 & 4.7 & 10.3 \\
 \hline
 Baseline       & 5.0 & 22.0 &  14.3 & 15.7 & 5.3 & 12.5  \\ 
 Ours & 3.8 & 17.5 & 13.8 & 12.7 & 5.2 & 10.5  \\ \hline
\end{tabular}}
\label{table:worldexpo}
\end{table}

\minisection{Evaluation on the UCSD dataset} The UCSD
dataset~\cite{chan2008privacy} has 2000 frames with a region of
interest (ROI) varying from 11 to 46 persons per image. The resolution
of each frame is fixed and small (238 $\times$ 158), so we change the
input of network to 112 $\times$ 112 by removing all layers after
pool4 in VGG-16. Thus the output retains the same 1/16 of input
size. We follow the same settings as~\cite{Sam_2017_CVPR}, using
frames between 600 and 1400 as training set, and the rest as test
set. In order to train our multi-task approach and compare fairly to
other methods, we generate ranked sets using the frames from 1 to 600
and test on the frames from 1401 to 2000 (and vice versa). We do not
collect additional unlabeled data since the training and test set are
from the same camera. Results are shown in Table~\ref{table:ucsd}.
Our baseline performs similarly to the state-of-the-art
  methods~\cite{Sam_2017_CVPR,li2018csrnet}, but slightly worse than
  ACSCP~\cite{shen2018crowd} and MCNN~\cite{zhang2016single}. Our
  multi-task approach reduces the baseline MAE from 1.60 to 1.17. Our
multi-task approach works on less dense datasets like UCSD because,
though the baseline might make incorrect predictions on patches with
the almost same number of headcounts, our learning-to-rank branch can
constrain it and ensure correct ranking predictions.

\begin{table}
\centering
\caption{MAE and MSE error on the UCSD dataset.}
\begin{tabular}{rcc}
\hline
\textbf{Method} & \textbf{MAE}   & \textbf{MSE}   \\ \hline \hline
Cross-scene~ \cite{zhang2015cross}       & 1.60 & 3.31 \\
MCNN~\cite{zhang2016single}      & 1.07 & \textbf{1.35} \\
Switching-CNN~\cite{Sam_2017_CVPR}     & 1.62 & 2.10 \\ 
ACSCP~\cite{shen2018crowd} & \textbf{1.04} & \textbf{1.35} \\ 
CSRNet~\cite{li2018csrnet} & 1.16 & 1.47 \\ \hline
 Baseline   & 1.60 & 2.13 \\
 Ours   & 1.17 &  1.55 \\ \hline
\end{tabular}

\label{table:ucsd}
\end{table}

\minisection{Evaluation on the UCF-QNRF dataset.}
UCF-QNRF~\cite{idrees2018composition} is a very challenging
  dataset consisting of 1,535 images with average of 815 people per
  image. The average resolution of images is much larger compared to
  other datasets, with images up to 6,000$\times$9,000 pixels. We
  resize all images to have a maximum dimension of 1024 pixels without
  changing the aspect ratio. The Keyword Dataset is used as unlabeled
  data to for the multi-task learning. Results are shown in
  Table~\ref{table:qnrf}, which indicate that our technique
  consistently improves counting performance -- even on datasets like
  UCF-QNRF with significantly more labeled training samples. Compared
  to our baseline, the MAE is reduced from 137 to 124 and MSE is down
  to 196. Our method outperforms all other methods including
  CompositionLoss~\cite{idrees2018composition} in MAE and performs
  slightly worse in MSE.

\begin{table}
\centering
\caption{MAE and MSE error on the UCF-QNRF dataset.}
\begin{tabular}{rcc}
\hline
\textbf{Method} & \textbf{MAE}   & \textbf{MSE}   \\ \hline \hline
MCNN~\cite{zhang2016single}      & 277 &  426 \\
Switching-CNN~\cite{Sam_2017_CVPR}     & 228 & 445 \\ 
CompositionLoss~\cite{idrees2018composition} &  132 & \textbf{191}  \\ \hline
 Baseline   & 137 & 228 \\
 Ours   & \textbf{124} & 196  \\ \hline
\end{tabular}

\label{table:qnrf}
\end{table}

\minisection{Evaluation on transfer learning.}
As proposed in~\cite{zhang2016single}, to demonstrate the
generalization of the learned model, we test our method in the
transfer learning setting by using Part\_A of the ShanghaiTech dataset
as the source domain and using UCF\_CC\_50 dataset as the target
domain. The model trained on Part\_A of ShanghaiTech is used to
predict the crowd scene images from UCF\_CC\_50 dataset, and the
results can be seen in Table~\ref{table:transfer}. Using only counting
information improves the MAE by 12\% compared to reported results
in~\cite{zhang2016single}. By combining both ranking and counting
datasets, the MAE decreases from 349.5 to 337.6, and MSE decreases
from 475.7 to 434.3. In conclusion, these results show that our method
significantly outperforms the only other work reporting results
on the task of cross-dataset crowd counting.

\begin{table}[tb]
\centering
\caption{Transfer learning across datasets. Models were trained on
  Part\_A of ShanghaiTech and tested on UCF\_CC\_50.}
\begin{tabular}{rcc}
\hline
\textbf{Method} & \textbf{MAE}   & \textbf{MSE}   \\ \hline \hline
MCNN~\cite{zhang2016single}      & 397.6 & 624.1 \\ \hline
 Counting only        & 349.5 & 475.7 \\
 Multi-task   & 337.6 & 434.3 \\ \hline
\end{tabular}

\label{table:transfer}
\end{table}

\subsubsection{Active learning for crowd counting}
\begin{figure}[tbp]
\centering
\includegraphics[width = 7.0cm]{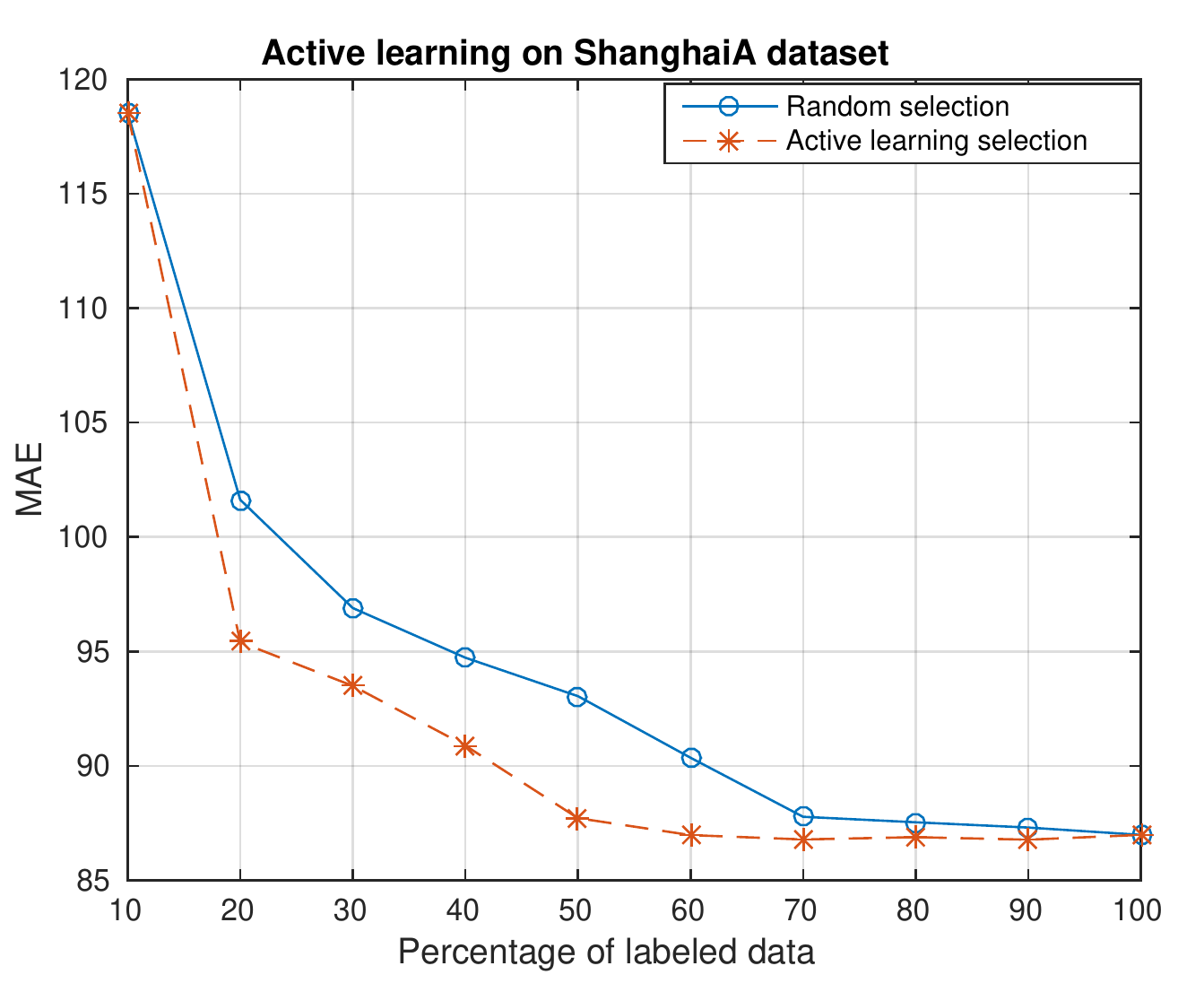}

\caption{Active learning results on Shanghai A. MAE is plotted as a function of the percentage of labeled data from the training set.} 
  \label{fig:active_count}
\end{figure}

We use the Shanghai Part\_A dataset to evaluate our active learning
approach on Crowd counting. We use $10\%$ of the training set as the
initially labeled training samples $D$ (and $E$ the remaining 90\%),
and set $S$ to add 10\% of the training images in each of the $T=9$
active learning cycles (see Algorithm~\ref{table:algorithm}). We use
$K=100$ in this experiment to evaluate ranking certainty. Again we
compare to the baseline of randomly selecting images from $E$. The
result is shown in Fig.~\ref{fig:active_count}. Our approach performs
consistently better than random selection in terms of MAE. We obtain
similar results with 20\% of the training data as random approach does
with 40\% -- reducing the labeling effort by 50\%. These results also
show that performance saturates after 60\% and no further improvement
is obtained by adding the last 40\% images considered least useful by
the active learning algorithm. The results clearly show that our
active learning method correctly identifies the images, which when
labeled, contribute most to an improved crowd counting network.     

\section{Conclusion}
\label{sec:conclusion}
In this article we explored ranking as a self-supervised proxy task for
regression problems. For many regression problems the collection of
supervised data is an expensive and laborious process. We showed, however, that
there exist some problems for which it is easy to obtain ranked image
sets, and that these ranked image sets can be exploited to improve the training
of the network. In addition, we proposed a method for fast
backpropagation for the ranking loss. This method removes the
redundant computation which is introduced by the multiple branches of
Siamese networks, and instead uses a single branch after which all
possible pairs of the minibatch are combined (rather than just a
selection of pairs).

We applied the proposed framework to two regression problems:
Image Quality Assessment and crowd counting. In the case of Image
Quality Assessment, the ranked data sets are formed by adding
increasing levels of distortions to images. For crowd counting,
ranked sets are formed by comparing image crops which are contained
within each other: a smaller image contained in another larger one will
contain the same number or fewer persons than the larger
image. Experimental results show that for both applications results
improve significantly when adding unlabeled data for the ranking
task. In addition, we have shown that the best results are obtained
when using our efficient backpropagation method in a multi-task
setting.

We have also shown that the proxy task can be used as an
informativeness measure for unlabeled images. The number of errors made
on the proxy task can be used to drive an active learning algorithm to
select the best images to label from a pool of unlabeled ones. These
images, once added to the training set, will most improve the
performance of the network. Experimental results show that,
  for both IQA and crowd counting, this method can reduce
  the labeling effort by a large margin.

\section*{Acknowledgments}
We acknowledge  the  Spanish project TIN2016-79717-R, the CHISTERA project M2CR (PCIN-2015-251) and the CERCA Programme / Generalitat de Catalunya. Xialei Liu acknowledges the Chinese Scholarship Council (CSC) grant No.201506290018. We also acknowledge the generous GPU donation from NVIDIA.

\ifCLASSOPTIONcaptionsoff
  \newpage
\fi

\bibliographystyle{IEEEtran}
\bibliography{refs}

%

\begin{IEEEbiography}[{\includegraphics[width=1in,height=1.25in,clip,keepaspectratio]{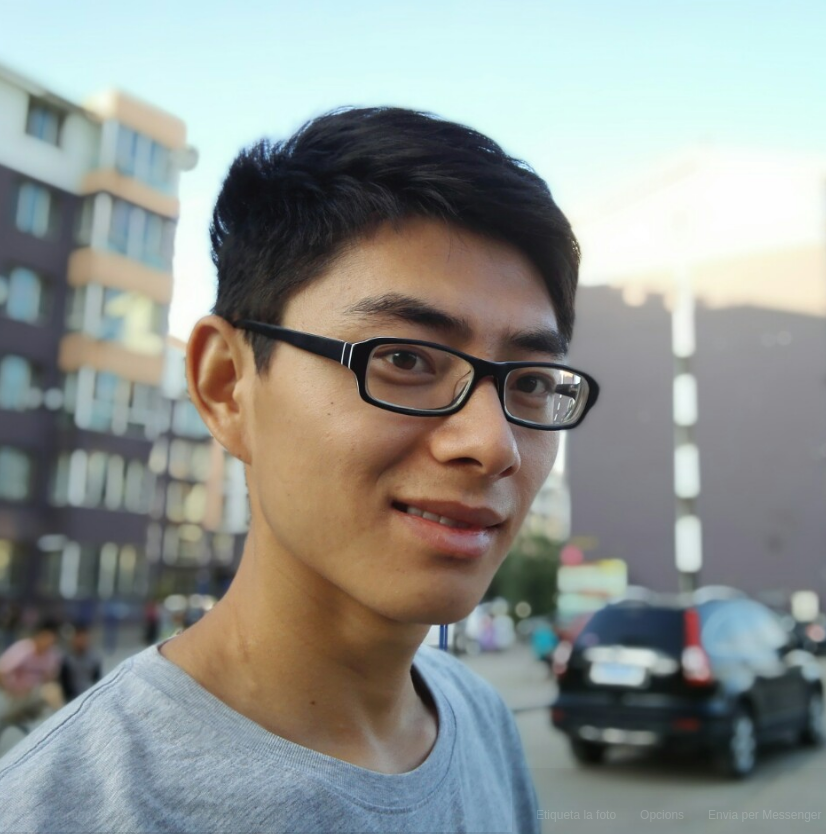}}]{Xialei Liu}
 Xialei Liu is a Ph.D. student in the Learning and Machine Perception (LAMP)
  group at the Universitat Autonoma de Barcelona. He received his
  B.S. degree in Information Engineering from Northwestern
  Polytechnical University, Xi'an, China, in 2013. He received an
  M.S. degree in Control Engineering from Northwestern Polytechnical
  University in March, 2016 and a second M.S. in Computer Vision from
  Universitat Autonoma de Barcelona in September, 2016. His research
  interests include image quality assessment, crowd counting,
  self-supervised learning, metric learning, lifelong learning and GANs.
\end{IEEEbiography}

\begin{IEEEbiography}[{\includegraphics[width=1in,height=1.25in,clip,keepaspectratio]{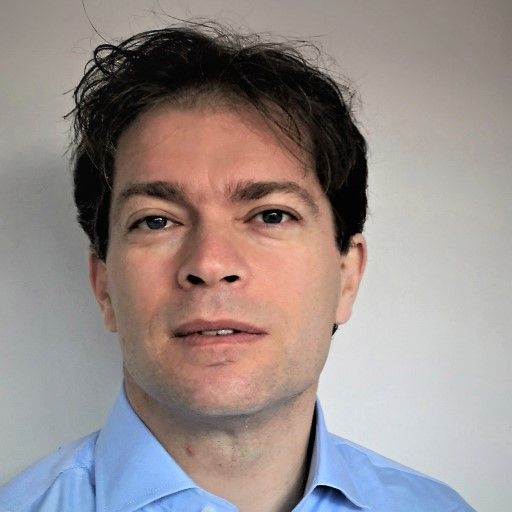}}]{Joost van de Weijer}
  Joost van de Weijer is a Senior Scientist at the Computer Vision Center Barcelona and leader of the LAMP team. He received his Ph.D. degree in 2005 from the University of Amsterdam. From 2005 to 2007, he was a Marie Curie Intra-European Fellow in the LEAR Team, INRIA Rhone-Alpes, France. From 2008 to 2012, he was a Ramon y Cajal Fellow at the Universidad Autonoma de Barcelona. His main research is on the machine learning applied to computer vision. 
\end{IEEEbiography}


\begin{IEEEbiography}[{\includegraphics[width=1in,height=1.25in,clip,keepaspectratio]{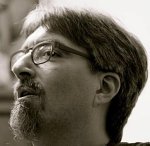}}]{Andrew D. Bagdanov}
  Andrew D. Bagdanov is Associate Professor at the University of
  Florence, Italy. He has held postdoctoral positions at the
  Universitat Autonoma de Barcelona, the Media Integration and
  Communication Center in Florence, Italy, and was a Ramon y Cajal
  Fellow at the Computer Vision Center, Barcelona. His research spans
  a broad spectrum of computer vision, image processing and machine
  learning.
\end{IEEEbiography}

\clearpage

\appendix[Distortions generated for TID2013]\label{sec:appendix}
The TID2013 dataset consists of 25 reference images with 3000
distorted images from 24 different distortion types at 5 degradation
levels. Mean Opinion Scores are in the range [0, 9]. Distortion types
include a range of noise, compression, and transmission artifacts. We
generate 17 out of the 24 distortions for training our networks. For
the distortions which we could not generate, we apply fine-tuning from
the network trained from the other ones. The generations details are
as follows (distortions in \textbf{bold} are synthetically generated,
while those in normal typeface we do not generate):

\begin{itemize}
\item \textbf{\#01 additive white Gaussian noise}: 
The local variance of the Gaussian noise added in RGB color space is set to be [0.001, 0.005, 0.01, 0.05].
\item \textbf{\#02 additive noise in color components}: 
The local variance of the Gaussian noise added in the YCbCr color space is set to be [0.0140, 0.0198, 0.0343, 0.0524].
\item \#03 additive Gaussian spatially correlated noise: there was
  insufficient detail in the original TID2013 paper~\cite{ponomarenko2013color} about how
  spatially correlated noise was generated and added to reference
  images.
\item \#04 masked noise: there was insufficient detail in the original
  TID2013 paper~\cite{ponomarenko2013color} about how masks were generated.
\item \textbf{\#05 high frequency noise}: The local variance of the
  Gaussian noise added in the Fourier domain is set to be [0.001,
  0.005, 0.01, 0.05] after which it is multiplied by a high-pass filter.
\item \textbf{\#06 impulse noise}: 
The local variance of ``salt \& pepper'' noise added in RGB color space is set to be [0.005, 0.01, 0.05, 0.1].
\item \textbf{\#07 quantization noise}: 
The quantization step is set to be [27, 39, 55, 76].
\item \textbf{\#08 Gaussian blur}: 2D circularly symmetric Gaussian
  blur kernels are applied with standard deviations set to be [1.2,
  2.5, 6.5, 15.2].
\item \textbf{\#09 image denoising}: The local variance of the
  Gaussian noise added in RGB color space is [0.001, 0.005, 0.01,
  0.05]. Followed by the same denoising process as in ~\cite{dabov2007image}. 
\item \textbf{\#10 JPEG compression}: 
The quality factor that determines the DCT quantization matrix is set to be [43, 12, 7, 4].
\item \textbf{\#11 JPEG2000 compression}: 
The compression ratio is set to be [52, 150, 343, 600].
\item \#12 JPEG transmission errors: the precise details of how JPEG
  transmission errors were introduced was not clear and we were
  unable to reproduce this distortion type.
\item \#13 JPEG2000 transmission errors: the precise details of how
  JPEG2000 transmission errors were introduced was not clear and we
  were unable to reproduce this distortion type.
\item \textbf{\#14 non eccentricity pattern noise}: Patches of size 15x15 are randomly moved to nearby regions~\cite{ponomarenko2013color}. The number of patches is set to [30, 70, 150, 300]. 
\item \textbf{\#15 local blockwise distortion of different intensity}:
Image patches of 32x32 are replaced by single color value (color block)~\cite{ponomarenko2013color}. The number of color blocks we distort is set to be [2, 4, 8,
  16]. 
\item \textbf{\#16 mean shift}: 
Mean value shifting generated in both directions is set to be: [-60,-45,-30,-15] and [15, 30, 45, 60]. 
\item \textbf{\#17 contrast change}: 
Contrast change generated in both directions is set to be:  [0.85, 0.7, 0.55, 0.4] and [1.2, 1.4, 1.6, 1.8].
\item \textbf{\#18 change of color saturation}: 
The control factor as in TID2013 paper~\cite{ponomarenko2013color} is set to be: [0.4, 0, -0.4, -0.8].
\item \textbf{\#19 multiplicative Gaussian noise}: 
The local variance of the Gaussian noise added is set to be [0.05, 0.09,0.13, 0.2].
\item \#20 comfort noise: the authors of~\cite{ponomarenko2013color} used a proprietary encoder unavailable to us.
\item \#21 lossy compression of noisy images: the authors of~\cite{ponomarenko2013color} used a proprietary encoder unavailable to us.
\item \textbf{\#22 image color quantization with dither}: 
The quantization step is set to be: [64, 32, 16, 8].
\item \textbf{\#23 chromatic aberrations}: 
The mutual shifting of in R and B channels is set to be [2, 6, 10, 14] and [1, 3, 5, 7], respectively.  
\item \#24 sparse sampling and reconstruction: the authors of~\cite{ponomarenko2013color} used a proprietary encoder unavailable to us.
\end{itemize}

\end{document}